\pdfoutput=1
\documentclass[10pt, logo, copyright]{nvidiatechreport}

\usepackage[authoryear,round]{natbib}

\definecolor{nvidiagreen}{RGB}{118,185,0}
\hypersetup{allcolors=nvidiagreen}
\makeatletter
\renewcommand{\abscontent}{%
    \par\vskip6pt
    {\noindent\bfseries\fontsize{12}{14}\selectfont Abstract\par}
    \vskip6pt
    \noindent
    \parbox{\dimexpr\linewidth}{\absfont \theabstract}
    \@ifundefined{@keywords}{}{
        \vskip1em \noindent \keywordsfont  Keywords: \@keywords}
}
\makeatother

\usepackage{amsmath,amsfonts,bm}

\def\eqref#1{equation~\ref{#1}}

\def\1{\bm{1}}

\DeclareMathAlphabet{\mathsfit}{\encodingdefault}{\sfdefault}{m}{sl}
\SetMathAlphabet{\mathsfit}{bold}{\encodingdefault}{\sfdefault}{bx}{n}

\usepackage{adjustbox}
\usepackage{url}
\usepackage{booktabs}
\usepackage{graphicx}
\usepackage{float}
\usepackage{xcolor}
\usepackage{listings}
\usepackage{needspace}
\usepackage{colortbl}
\usepackage{multirow}
\usepackage{pgfplots}
\pgfplotsset{compat=1.18}
\usepgfplotslibrary{groupplots}
\usepackage{capt-of}
\definecolor{gaincolor}{HTML}{2E7D32}
\definecolor{losscolor}{HTML}{C62828}
\definecolor{oursframe}{HTML}{76B900}
\definecolor{bestours}{HTML}{E3F1C8}
\definecolor{bestother}{HTML}{E4E3DE}
\definecolor{rowsep}{HTML}{E1E0D9}
\definecolor{stdretr}{HTML}{8F8E89}
\definecolor{basecurve}{HTML}{9F9E99}
\definecolor{ourstext}{HTML}{4F7D00}
\definecolor{basetext}{HTML}{6F6E69}
\definecolor{axisgray}{HTML}{C3C2B7}
\definecolor{ourslight}{HTML}{B5D86F}
\definecolor{bandblue}{HTML}{EDF3FA}
\newcommand{\relup}[1]{{\scriptsize\textcolor{gaincolor}{+#1\%}}}
\newcommand{\reldown}[1]{{\scriptsize\textcolor{losscolor}{$-$#1\%}}}
\newcommand{\relzero}{{\scriptsize 0.0\%}}
\newcolumntype{O}{!{\color{oursframe}\vrule width 1.2pt}c!{\color{oursframe}\vrule width 1.2pt}}
\newcommand{\restablescale}{0.735}  
\lstdefinestyle{paperprompt}{
  basicstyle=\ttfamily\footnotesize,
  columns=fullflexible,
  keepspaces=true,
  showstringspaces=false,
  breaklines=true,
  breakatwhitespace=true,
  breakautoindent=false,
  breakindent=0pt,
  frame=single,
  framerule=0.3pt,
  rulecolor=\color{black!25},
  backgroundcolor=\color{black!2},
  framesep=4pt,
  xleftmargin=5pt,
  xrightmargin=5pt,
  aboveskip=3pt,
  belowskip=4pt
}

\title{Frozen Models, Evolving Expertise:\\
Model-Agnostic Learning from Deployment Experience for Multimodal Medical AI}


\author[1]{Yexiao He}
\author[2]{Yucheng Tang}
\author[2]{Pengfei Guo}
\author[2]{Yufan He}
\author[2]{Andriy Myronenko}
\author[2]{Can Zhao}
\author[1]{Ang Li}
\author[2]{Daguang Xu}
\author[2]{Dong Yang}

\medskip

\affil[1]{University of Maryland, College Park}
\affil[2]{NVIDIA}

\correspondingauthor{Dong Yang (dongy@nvidia.com)}

\makeatletter
\let\nvfirststyle@orig\ps@firststyle
\def\ps@firststyle{\nvfirststyle@orig
  \fancyfoot[L]{\footerfont\itshape{Yexiao He: yexiaohe@umd.edu; Corresponding author: \the\correspondingauthor\\}%
    \ifthenelse{\boolean{copyright}}{\copyrightext}{}}}
\makeatother

\begin{abstract}
Large language models (LLMs) and vision-language models (VLMs) are usually frozen after deployment, so they do not learn from the cases they solve. This is especially concerning in medicine, where new clinical evidence, updated guidelines, and new therapies can change established practice. Fine-tuning can update the model, but it requires access to model weights and additional training. Parameter-free methods avoid training, but they may overfit a fixed validation set, lack reliable domain knowledge, or lose visual details by saving experience only as text. To address these limitations, we present a model-agnostic framework that allows frozen LLMs and VLMs to learn from deployment experience through three forms of external expertise: a \emph{Skill} that guides reasoning and tool use, a \emph{Knowledge Memory} that stores reliable facts supported by earlier cases or trusted external evidence, and a \emph{Multimodal Knowledge Base} that keeps visual examples and guides the model to relate each retrieved case to the current image. Instead of relying on a fixed validation set, a validation strategy keeps an update only if it helps on new cases without degrading performance on earlier ones. Across six benchmarks covering clinical diagnosis, clinical workflows, medical reasoning, and medical and non-medical visual reasoning, and with four open-weight and closed-source base models, our framework improves performance during online deployment by up to 34.2\% over the base model on medical tasks, generalizes to unseen cases, transfers to other models without further optimization, and works in non-medical domains.
\end{abstract}

\begin{document}

\maketitle

\section{Introduction}
\label{sec:introduction}

Large language models (LLMs) and vision-language models (VLMs) can now solve complex problems, use external tools, and reason over both text and images~\citep{yao2022react,liu2023visual}. However, once these models are deployed, their parameters are usually frozen, so their internal knowledge and capabilities do not automatically improve with new experience~\citep{zhao2024expel,majumder2023clin}. This limitation is especially problematic in medicine, where new clinical evidence, updated guidelines, and new therapies and drugs can change established clinical practice~\citep{shojania2007quickly}. During deployment, a model can generate experience: it solves cases, searches for evidence, makes mistakes, and receives feedback. A verified clinical case may reveal a better diagnostic procedure, a reliable medical fact, or a useful visual pattern. If a model cannot reuse this experience, it has to start from scratch whenever it sees a similar case and may make the same mistakes again~\citep{shinn2023reflexion,zhao2024expel}. This raises an important question: \emph{how can a frozen model learn from deployment experience and develop stronger capabilities in a specific domain or task like medicine?}

Recent work has explored how models can learn from interactions and feedback, either by saving past episodes and reusable procedures outside the model~\citep{shinn2023reflexion,zhao2024expel,majumder2023clin,alzubi2026evoskill,yang2026skillopt,zhang2026agentic} or by updating model parameters through fine-tuning or reinforcement learning on interaction trajectories~\citep{song2024agentbank,yuan2025agent}.

However, improving a deployed model in these ways has practical limitations. Training requires access to the model weights, which may be impossible for closed-source models~\citep{zhao2024expel}. Training also incurs additional costs and depends on the quality of training data~\citep{shinn2023reflexion,song2024agentbank,yuan2025agent}. Parameter-free methods such as EvoSkill and SkillOpt avoid retraining, but they use a fixed validation set to select the updates to their skill documents~\citep{alzubi2026evoskill,yang2026skillopt}. Repeatedly selecting updates on the same validation cases can lead to overfitting and hinder generalization to new cases~\citep{dwork2015reusable}. Moreover, experience derived only from interaction trajectories may lack the reliable domain knowledge needed for new cases. Finally, methods such as Reflexion, ExpeL, and ACE save learned experience as text~\citep{shinn2023reflexion,zhao2024expel,zhang2026agentic}, which can omit useful visual details.

To address these limitations, we develop an experience-driven learning framework for medical AI. It improves frozen LLMs and VLMs by accumulating and updating external expertise. The framework maintains three components. The \emph{Skill} provides reusable guidance for reasoning and tool use. It is periodically revised based on which reasoning and tool-use steps led to high or low task scores on earlier cases. A \emph{Knowledge Memory} stores reliable facts supported by earlier cases or trusted external evidence. For multimodal tasks, a \emph{Multimodal Knowledge Base (MMKB)} saves earlier visual cases so that the original visual evidence remains available to later reasoning. These three components address general needs in medicine: reusing better diagnostic procedures, reliable medical facts, and useful visual patterns. To avoid overfitting a fixed validation set, we also design a validation strategy that tests each update to the learned expertise on new and earlier cases to keep only updates that generalize to new cases without degrading performance on earlier ones. In this way, the system can improve during deployment while the model parameters remain unchanged.

We evaluate the framework on six benchmarks covering interactive clinical diagnosis, clinical workflows, text-based medical reasoning, and medical and non-medical visual reasoning~\citep{schmidgall2024agentclinic,liu2026medchain,zhou2025automating,yao2026medical,xu2504visulogic,cao2026qcaleval}. Our framework improves online performance over the base model by up to 34.2\% on medical tasks, generalizes to unseen cases, and transfers to other models without further optimization. These results show that deployment experience can produce reusable expertise.

Our main contributions are as follows:
\begin{itemize}
    \item \textbf{Continual improvement through external expertise.} We develop a framework that enables frozen LLMs and VLMs to continually improve during deployment without training by turning experience into reusable external expertise. Because it does not require access to model weights, it applies to both open-weight and closed-source models.

    \item \textbf{Persistent multimodal expertise.}
    We introduce MMKB, which preserves images, source answers, and available explanations from reference data and completed cases. It guides the model to examine the evidence behind each reference answer and to compare relevant observations and conditions with the current case, so that visual patterns learned from previous cases can improve reasoning on new images. Used together with the Skill and Knowledge Memory, MMKB further improves performance on visual tasks.

    \item \textbf{Reusable expertise across models and domains.} Expertise learned with one model can be reused by other models without further optimization, so it does not need to be relearned when the base model changes. The framework also improves non-medical visual reasoning tasks, showing that it is not limited to medicine.
\end{itemize}

\section{Related Work}
\label{sec:related_work}

\paragraph{External memory and reusable procedures.}
Parameter-free methods preserve expertise outside the base model. Many of them accumulate reusable experience from earlier trajectories, such as reflections, strategies, guidance, or executable skills, and add it directly~\citep{shinn2023reflexion,zhao2024expel,majumder2023clin,suzgun2026dynamic,ouyang2026reasoningbank,li2024agent,fu2024autoguide,chen2024automanual,wang2024agent,wang2023voyager,zheng2025skillweaver,ni2603trace2skill,zhang2026agentic}. Others, such as GEPA, EvoSkill, and SkillOpt, generate candidate updates and select them by their performance on a held-out validation set~\citep{agrawal2025gepa,alzubi2026evoskill,yang2026skillopt}. However, experience added without checking its effect may not help new cases, repeatedly optimizing and validating on the same fixed set of cases can overfit that set, and memories and procedures derived only from interaction trajectories may lack reliable domain knowledge. Our framework instead tests each candidate update on new cases and a representative buffer of past cases, and Knowledge Memory adds supported facts from verified cases or trusted external sources.

\paragraph{Fine-tuning and soft prompting.}
Many methods update model parameters on interaction trajectories through supervised fine-tuning~\citep{chen2023fireact,zeng2310agenttuning,song2024agentbank} or reinforcement learning~\citep{yuan2025agent,luo2508agent}. These methods require access to model weights, incur substantial training costs, typically rely on many reliable trajectories, and must be repeated for each base model. \citet{gare2026your} keep the VLM backbone frozen and train a few continuous prompt tokens, which can transfer to a newer model version, but training them still requires gradients through the model weights, so the method applies only to open-weight models. Our framework instead updates external expertise from deployment experience without gradient access, so it also applies to closed-source models, and the learned expertise can transfer to other models without further optimization.

\paragraph{Multimodal memory and retrieval.}
Most parameter-free experience learning methods return learned experience to the model as text. MIRIX and M3-Agent instead retain multimodal observations for long-term recall~\citep{wang2507mirix,long2026seeing}, while Med-RwR, RULE, and MMed-RAG retrieve textual or visual evidence for medical VLMs~\citep{wang2025proactive,xia2024rule,xia2410mmed}. However, these systems mainly provide the retrieved content as context, without explicit instructions to analyze it or compare it with the current case. MMKB instead guides the model to examine how each retrieved visual case, with its known answer, relates to the current image.

\section{Method}
\label{sec:method}

\subsection{Framework Overview}
\label{sec:method_overview}

\begin{figure}[t]
    \centering
    \includegraphics[width=0.88\linewidth]{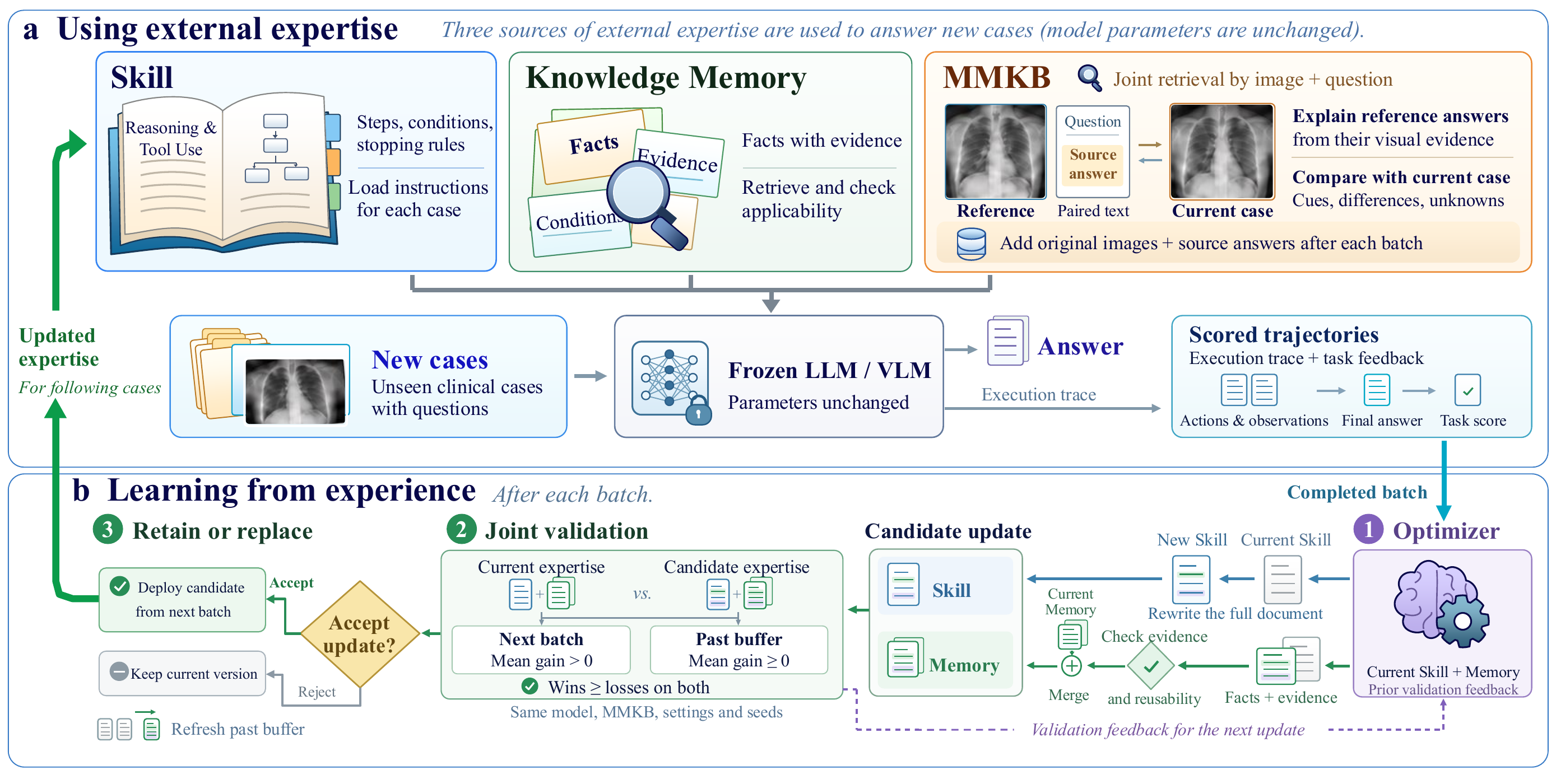}
    \caption{\textbf{Framework overview.} (a) A frozen model uses Skill, Knowledge Memory, and MMKB to answer new cases. MMKB supports comparison with visual references and their source answers. (b) The optimizer proposes new Skills and supported knowledge additions from scored trajectories. The candidate update is tested on new and past cases before deployment, and validation feedback guides the next update.}
    \label{fig:method_overview}
\end{figure}

Our goal is to let a frozen model improve at a task by learning from verified deployment experience. As shown in Figure~\ref{fig:method_overview}, the framework maintains a \emph{Skill} for reasoning and tool-use procedures, a \emph{Knowledge Memory} for reusable facts, and an \emph{MMKB} for visual examples. Together, they provide guidance on how to solve a task and evidence that may be missing from the model's own knowledge.

\paragraph{Problem setting.}
A frozen base model processes a stream of cases in batches $F_1,F_2,\ldots$, where $F_t$ denotes the $t$-th batch. Let $\mathcal{E}_t=(S_t,K_t,V_t)$ denote the Skill, Knowledge Memory, and MMKB used for batch $F_t$, and let $s_i(\mathcal{E}_t)$ denote the task score of case $i$ when the base model uses $\mathcal{E}_t$. The model answers every case in $F_t$ before the task scores and source answers of these cases are released, so this feedback can affect only later batches. The source answer of a case is the reference answer or report provided by its dataset, and we call a case verified once its source answer is released. $S_1$ and $K_1$ are empty, and for visual tasks $V_1$ contains the initial reference data. The goal is to update $\mathcal{E}_t$ so that later cases receive higher scores, while the model parameters remain fixed.

The base model receives the Skill $S_t$ in its task instructions, queries Knowledge Memory $K_t$ when needed, and retrieves MMKB references from $V_t$ for visual tasks. After each batch, an LLM optimizer proposes a new Skill and knowledge additions from scored trajectories, and these updates are validated before deployment (Section~\ref{sec:joint_validation}). The MMKB $V_t$ grows from completed visual cases with source answers.

\subsection{Learning Reusable Procedures}
\label{sec:skill_evolution}

The Skill is a natural-language document that describes useful steps, their conditions of use, and when additional actions are unnecessary. For example, a diagnostic Skill may suggest checking evidence that distinguishes competing diagnoses before ordering another test. The complete document is available from the start of each case, and the model applies it according to current evidence and task requirements.

After each batch, the optimizer reviews the current Skill and Memory, the scored trajectories, and the comparison results of the previous candidate (Section~\ref{sec:joint_validation}). Based on successful and unsuccessful actions, it rewrites the whole Skill document: it adds missing steps, revises overly broad rules, removes unhelpful guidance, and keeps the Skill within a length limit. The optimizer rewrites instead of appending because appending makes the Skill, and thus the model's context, longer after every batch and can leave conflicting rules; both can degrade performance. For example, if one batch suggests ``order a chest X-ray for patients with cough and fever'' and a later batch shows that the X-ray is often unnecessary when vital signs and lung examination are normal, rewriting merges them into one rule: ``order a chest X-ray for patients with cough and fever when vital signs or lung examination are abnormal.'' The optimizer is instructed not to include case identifiers or answer keys, and the proposed Skill is validated before it is used on subsequent cases.

\subsection{Acquiring and Reusing Knowledge}
\label{sec:knowledge_memory}

Even with a better procedure, the model may lack facts needed for a decision. Knowledge Memory stores facts and relations with supporting evidence and conditions of use. When the model asks a knowledge question through a tool call, text retrieval finds potentially relevant entries, and a separate model checks whether they directly answer the question and apply to the case. If none is suitable, the system searches trusted external sources, such as PubMed. 

Knowledge candidates come from these searches and from the optimizer's analysis of completed cases. Case-based proposals must cite supporting trajectories and quote their evidence. The model checks that the evidence exists, supports the fact, and justifies reuse beyond the original case. Multiple supporting cases are normally required; a single case is sufficient only with direct support from a trusted external source. Patient-specific observations and individual answer keys are excluded.
Supported facts are merged with the current Memory to form a candidate update. Factual support alone does not establish that an addition will improve decisions, so we test the updated expertise before deployment.

\subsection{Learning from Visual References}
\label{sec:mmkb}

Textual lessons may omit visual details. MMKB retains original images, their questions or context, source answers, and explanations when available. It is initialized from reference cases outside the evaluated stream (Table~\ref{tab:mmkb_sources}) and grows through completed deployment cases.

A fixed multimodal encoder jointly represents each case's images and available question text. Cosine similarity retrieves relevant references.

MMKB guides how retrieved references are used, rather than only which references are retrieved. For each reference, the model examines how its images and context support the source answer, then identifies which observations and conditions also apply to the current case. It must consider differences, missing evidence, and contradictions before answering. This comparison keeps the current decision grounded in the current input while drawing on experience from earlier cases.

After batch $F_t$, all cases in $F_t$ that have images and source answers are added to $V_t$ with their source answers and existing explanations, not the model's predictions, to form $V_{t+1}$. Even an incorrectly answered case can therefore become a useful reference. These references are retained even if proposed updates to Skill and Memory are rejected or earlier versions are restored. New references are available from the following batch. Retrieval also excludes the current case, cases from its source group, and duplicate images (Appendix~\ref{sec:appendix_method}).

\subsection{Validating Updates During Deployment}
\label{sec:joint_validation}

To test whether an update helps beyond the cases that produced it, we compare the current expertise $\mathcal{E}_t$ with a candidate version $\widetilde{\mathcal{E}}_t=(\widetilde{S}_t,\widetilde{K}_t,V_t)$ that includes Skill and Memory updates proposed after batch $F_{t-1}$, where $\widetilde{S}_t$ and $\widetilde{K}_t$ denote the candidate Skill and Knowledge Memory. Both versions are tested on $F_t$ and a representative buffer of past cases $B_t$. The buffer is initialized with cases from $F_1$, so the first comparison takes place at $t=2$. The base model first answers $F_t$ using the current expertise, and these answers are final; the candidate's answers are used only for the comparison.

Both versions are evaluated under the same base model, MMKB $V_t$, task settings, and random seed for each case. When a past case from $B_t$ is replayed, its own MMKB entry, including its images and source answer, is excluded from retrieval. For $D\in\{F_t,B_t\}$, define
\begin{equation}
    \overline{\Delta}_{D}=\frac{1}{|D|}\sum_{i\in D}
    \left[s_i(\widetilde{\mathcal{E}}_t)-s_i(\mathcal{E}_t)\right],
\end{equation}
where $s_i$ is the task score. Let $W_D$ and $L_D$ count cases on which the candidate scores higher and lower. We accept the update only if
\begin{equation}
    \overline{\Delta}_{F_t}>0,\quad W_{F_t}\geq L_{F_t},\qquad\qquad \overline{\Delta}_{B_t}\geq 0,\quad W_{B_t}\geq L_{B_t}.
\label{eq:joint_acceptance}
\end{equation}
New cases must show improvement, while past cases may tie but not decline on average. For tasks with binary scores, the win--loss condition follows from the mean condition. For tasks with continuous scores, it additionally rejects a candidate whose mean gain comes from large improvements on a few cases while more cases decline. If the update is accepted, $\mathcal{E}_{t+1}=(\widetilde{S}_t,\widetilde{K}_t,V_{t+1})$.

After each comparison, an LLM refreshes the buffer using only task goals and question texts, not answers or scores. It replaces some buffer cases with cases from $F_t$ so that the buffer covers the range of questions seen so far. Periodically, the current version is also compared on the buffer with an earlier retained version, and the earlier Skill and Memory are restored if the current version performs worse. These checks reduce the risk of overfitting to reused validation cases.

The optimizer then uses trajectories generated with the selected version on the completed batch to propose the next update. Appendices~\ref{sec:appendix_method} and~\ref{sec:appendix_prompts} give the key settings and prompt templates.

\section{Experiments}
\label{sec:experiments}

We test whether deployment experience improves a frozen model, whether each expertise component contributes to that improvement, and whether the learned expertise remains useful on unseen cases or with a different base model. These comparisons distinguish reusable expertise from corrections that help only the cases used for learning.

\subsection{Experimental Setup}
\label{sec:experimental_setup}

\paragraph{Benchmarks and models.}
We evaluate six benchmarks. AgentClinic-MedQA~\citep{schmidgall2024agentclinic}, MedChain~\citep{liu2026medchain}, and MedThink-Bench~\citep{zhou2025automating} cover interactive diagnosis, multi-stage clinical workflows, and text-based medical reasoning. MedThinkVQA~\citep{yao2026medical} adds multi-image diagnosis. VisuLogic~\citep{xu2504visulogic} and QCalEval~\citep{cao2026qcaleval} test visual logic and quantum-calibration reasoning beyond medicine. We use Qwen3.8-27B, Nemotron-3-Nano-Omni-30B, GPT-5.6-Sol, and GPT-4o-mini as frozen base models, with GPT-5.6-Sol as the optimizer. MMKB is enabled only for image tasks.

\paragraph{Comparisons.}
Base retains each benchmark's native task interface and external knowledge searches, without Skill, Knowledge Memory, or MMKB. We also compare with ACE~\citep{zhang2026agentic}, which maintains an evolving context, and SkillOpt~\citep{yang2026skillopt}, which learns reusable Skills. ACE is included in the online and frozen comparisons; SkillOpt is included in the frozen comparison. Task inputs and scoring rules are matched within each comparison.

\paragraph{Online protocol.}
Each benchmark follows a fixed case order with seed 42, and our framework updates after batches of 32 cases. All predictions are recorded before their feedback is used for learning. Appendix~\ref{sec:appendix_method} gives dataset sizes, retrieval settings, and input limits.

\paragraph{Evaluation.}
We report diagnosis accuracy (\%) for AgentClinic-MedQA; answer accuracy (\%) for MedThink-Bench, MedThinkVQA, and VisuLogic; the official overall score for QCalEval; and MedChain's Average on a 0--1 scale. MedChain's Average covers six metrics across its five tasks~\citep{liu2026medchain}; Appendix~\ref{sec:appendix_method} gives the definition. The online score uses only the deployed model's original predictions; candidate evaluations and replay do not contribute. Frozen and transfer evaluations retain local expertise retrieval but disable optimization, writes, and new external searches. Tables~\ref{tab:online_results} and~\ref{tab:frozen_results} report these scores and their relative change from Base.

\subsection{Online Improvement During Deployment}
\label{sec:online_improvement}

\begin{figure}[t]
\centering
\begin{tikzpicture}
\begin{groupplot}[
  group style={group size=5 by 2, horizontal sep=0.8cm, vertical sep=1.15cm},
  scale only axis, width=2.03cm, height=1.6cm,
  axis x line*=bottom, axis y line*=left,
  x axis line style={draw=none}, y axis line style={draw=none},
  xtick style={draw=axisgray, line width=0.5pt}, ytick style={draw=none},
  tick align=outside, major tick length=2pt,
  ymajorgrids, grid style={draw=rowsep, line width=0.5pt},
  ticklabel style={font=\scriptsize, text=black!50},
  xticklabel style={/pgf/number format/1000 sep={}},
  yticklabel style={/pgf/number format/fixed},
  title style={at={(0,1)}, anchor=base west, font=\footnotesize, yshift=2pt},
  clip mode=individual, every axis plot/.append style={line join=round},
  legend style={draw=none, fill=none, font=\footnotesize},
  legend image code/.code={\draw[#1] (0cm,0cm) -- (0.6cm,0cm);},
]
\nextgroupplot[title={AgentClinic}, xmin=0, xmax=226.8, ymin=73.95, ymax=84.8, xtick={0,100,200}, ytick={75,78,81,84}]
\addplot[fill=oursframe, fill opacity=0.13, draw=none, forget plot] coordinates {(32,75) (64,76.56) (96,79.17) (128,82.03) (160,83.75) (192,82.81) (214,83.64) (214,78.97) (192,78.65) (160,80.62) (128,80.47) (96,79.17) (64,76.56) (32,75)} -- cycle;
\addplot[basecurve, line width=1.2pt] coordinates {(32,75) (64,76.56) (96,79.17) (128,80.47) (160,80.62) (192,78.65) (214,78.97)};
\addplot[oursframe, line width=1.7pt] coordinates {(32,75) (64,76.56) (96,79.17) (128,82.03) (160,83.75) (192,82.81) (214,83.64)};
\addplot[basecurve, only marks, mark=*, mark size=1.3pt, forget plot] coordinates {(214,78.97)};
\addplot[oursframe, only marks, mark=*, mark size=1.5pt, forget plot] coordinates {(214,83.64)};
\draw[axisgray, line width=0.5pt] (axis cs:0,73.95) -- (axis cs:214,73.95);
\nextgroupplot[title={MedThink-Bench}, xmin=0, xmax=530, ymin=64.58, ymax=75.42, xtick={0,250,500}, ytick={66,69,72,75}]
\addplot[fill=oursframe, fill opacity=0.13, draw=none, forget plot] coordinates {(131.17,72.12) (160,74.38) (192,73.44) (224,73.21) (256,73.05) (288,73.26) (320,72.19) (352,71.88) (384,71.88) (416,71.63) (448,71.65) (480,71.46) (500,71.8) (500,68.4) (480,67.71) (448,66.21) (416,66.08) (384,67.18) (352,67.58) (320,68.75) (288,69.65) (256,68.61) (224,70.89) (192,71.62) (160,71.25) (131.17,72.12)} -- cycle;
\addplot[basecurve, line width=1.2pt] coordinates {(32,65.62) (64,68.75) (96,70.83) (128,72.22) (160,71.25) (192,71.62) (224,70.89) (256,68.61) (288,69.65) (320,68.75) (352,67.58) (384,67.18) (416,66.08) (448,66.21) (480,67.71) (500,68.4)};
\addplot[oursframe, line width=1.7pt] coordinates {(32,65.62) (64,67.19) (96,68.75) (128,71.88) (160,74.38) (192,73.44) (224,73.21) (256,73.05) (288,73.26) (320,72.19) (352,71.88) (384,71.88) (416,71.63) (448,71.65) (480,71.46) (500,71.8)};
\addplot[basecurve, only marks, mark=*, mark size=1.3pt, forget plot] coordinates {(500,68.4)};
\addplot[oursframe, only marks, mark=*, mark size=1.5pt, forget plot] coordinates {(500,71.8)};
\draw[axisgray, line width=0.5pt] (axis cs:0,64.58) -- (axis cs:500,64.58);
\nextgroupplot[title={MedThinkVQA}, xmin=0, xmax=763.2, ymin=51.12, ymax=72.6, xtick={0,250,500}, ytick={54,60,66,72}, legend style={at={(0.5,1.3)}, anchor=south, legend columns=2, /tikz/every even column/.append style={column sep=10pt}}]
\addplot[fill=oursframe, fill opacity=0.13, draw=none, forget plot] coordinates {(283.57,63.45) (288,63.54) (297.35,63.79) (297.35,63.79) (288,63.36) (283.57,63.45)} -- cycle;
\addplot[fill=oursframe, fill opacity=0.13, draw=none, forget plot] coordinates {(357.12,65.51) (384,66.41) (416,67.79) (448,68.53) (479.1,68.54) (479.1,68.54) (448,67.09) (416,66.95) (384,65.75) (357.12,65.51)} -- cycle;
\addplot[fill=oursframe, fill opacity=0.13, draw=none, forget plot] coordinates {(500.45,68.55) (512,68.55) (514.42,68.56) (514.42,68.56) (512,68.53) (500.45,68.55)} -- cycle;
\addplot[fill=oursframe, fill opacity=0.13, draw=none, forget plot] coordinates {(605.97,68.12) (608,68.09) (622.64,68.39) (622.64,68.39) (608,68.07) (605.97,68.12)} -- cycle;
\addplot[fill=oursframe, fill opacity=0.13, draw=none, forget plot] coordinates {(683.26,68.9) (704,68.89) (720,68.89) (720,68.32) (704,68.74) (683.26,68.9)} -- cycle;
\addplot[basecurve, line width=1.2pt] coordinates {(32,59.38) (64,62.5) (96,69.79) (128,68.09) (160,67.38) (192,66.27) (224,65.96) (256,64.02) (288,63.36) (320,64.81) (352,65.47) (384,65.75) (416,66.95) (448,67.09) (480,68.58) (512,68.53) (544,68.85) (576,68.97) (608,68.07) (640,68.78) (672,68.98) (704,68.74) (720,68.32)};
\addplot[oursframe, line width=1.7pt] coordinates {(32,53.12) (64,57.81) (96,64.58) (128,65.62) (160,63.75) (192,62.5) (224,62.05) (256,62.89) (288,63.54) (320,64.38) (352,65.34) (384,66.41) (416,67.79) (448,68.53) (480,68.54) (512,68.55) (544,68.57) (576,68.58) (608,68.09) (640,68.75) (672,68.9) (704,68.89) (720,68.89)};
\addplot[basecurve, only marks, mark=*, mark size=1.3pt, forget plot] coordinates {(720,68.32)};
\addplot[oursframe, only marks, mark=*, mark size=1.5pt, forget plot] coordinates {(720,68.89)};
\draw[axisgray, line width=0.5pt] (axis cs:0,51.12) -- (axis cs:720,51.12);
\legend{Base, Ours}
\nextgroupplot[title={VisuLogic}, xmin=0, xmax=1060, ymin=36.52, ymax=58.36, xtick={0,500,1000}, ytick={42,48,54}]
\addplot[fill=oursframe, fill opacity=0.13, draw=none, forget plot] coordinates {(53.33,47.92) (64,48.44) (96,44.79) (128,43.75) (160,46.25) (192,44.27) (224,46.43) (256,47.66) (288,46.88) (320,45.62) (352,45.74) (384,46.09) (416,48.32) (448,47.77) (480,48.54) (512,48.63) (544,48.16) (576,48.61) (608,49.34) (640,49.69) (672,50) (704,49.86) (736,50.27) (768,50.13) (800,49.88) (832,50.36) (864,51.04) (896,50.67) (928,49.89) (960,50) (992,50.1) (1000,50) (1000,45.8) (992,46.07) (960,45.94) (928,45.69) (896,45.76) (864,45.72) (832,44.59) (800,44.62) (768,44.53) (736,44.43) (704,44.32) (672,43.75) (640,43.59) (608,42.93) (576,42.36) (544,42.28) (512,42.97) (480,42.71) (448,42.19) (416,42.07) (384,40.36) (352,38.64) (320,39.38) (288,40.62) (256,41.8) (224,44.2) (192,43.75) (160,43.12) (128,41.41) (96,39.58) (64,43.75) (53.33,47.92)} -- cycle;
\addplot[basecurve, line width=1.2pt] coordinates {(32,56.25) (64,43.75) (96,39.58) (128,41.41) (160,43.12) (192,43.75) (224,44.2) (256,41.8) (288,40.62) (320,39.38) (352,38.64) (384,40.36) (416,42.07) (448,42.19) (480,42.71) (512,42.97) (544,42.28) (576,42.36) (608,42.93) (640,43.59) (672,43.75) (704,44.32) (736,44.43) (768,44.53) (800,44.62) (832,44.59) (864,45.72) (896,45.76) (928,45.69) (960,45.94) (992,46.07) (1000,45.8)};
\addplot[oursframe, line width=1.7pt] coordinates {(32,46.88) (64,48.44) (96,44.79) (128,43.75) (160,46.25) (192,44.27) (224,46.43) (256,47.66) (288,46.88) (320,45.62) (352,45.74) (384,46.09) (416,48.32) (448,47.77) (480,48.54) (512,48.63) (544,48.16) (576,48.61) (608,49.34) (640,49.69) (672,50) (704,49.86) (736,50.27) (768,50.13) (800,49.88) (832,50.36) (864,51.04) (896,50.67) (928,49.89) (960,50) (992,50.1) (1000,50)};
\addplot[basecurve, only marks, mark=*, mark size=1.3pt, forget plot] coordinates {(1000,45.8)};
\addplot[oursframe, only marks, mark=*, mark size=1.5pt, forget plot] coordinates {(1000,50)};
\draw[axisgray, line width=0.5pt] (axis cs:0,36.52) -- (axis cs:1000,36.52);
\nextgroupplot[title={QCalEval}, xmin=0, xmax=257.6, ymin=70.63, ymax=85.62, xtick={0,100,200}, ytick={72,76,80,84}]
\addplot[fill=oursframe, fill opacity=0.13, draw=none, forget plot] coordinates {(32,76.01) (64,76.79) (96,79.32) (128,79.84) (160,81.55) (192,82.16) (224,83.33) (243,84.17) (243,72.29) (224,72.77) (192,73.53) (160,73.62) (128,73.08) (96,73.43) (64,72.08) (32,73.31)} -- cycle;
\addplot[basecurve, line width=1.2pt] coordinates {(32,73.31) (64,72.08) (96,73.43) (128,73.08) (160,73.62) (192,73.53) (224,72.77) (243,72.29)};
\addplot[oursframe, line width=1.7pt] coordinates {(32,76.01) (64,76.79) (96,79.32) (128,79.84) (160,81.55) (192,82.16) (224,83.33) (243,84.17)};
\addplot[basecurve, only marks, mark=*, mark size=1.3pt, forget plot] coordinates {(243,72.29)};
\addplot[oursframe, only marks, mark=*, mark size=1.5pt, forget plot] coordinates {(243,84.17)};
\draw[axisgray, line width=0.5pt] (axis cs:0,70.63) -- (axis cs:243,70.63);
\nextgroupplot[title={MedChain Task 1}, xmin=0, xmax=2503.7, ymin=44.5, ymax=69, xtick={0,1000,2000}, ytick={48,56,64}]
\addplot[fill=oursframe, fill opacity=0.13, draw=none, forget plot] coordinates {(32,53.12) (64,60.94) (96,61.46) (128,64.84) (160,65) (192,64.58) (224,64.73) (256,64.45) (384,64.58) (512,63.87) (640,64.22) (768,65.76) (896,66.63) (1024,66.6) (1152,66.15) (1280,66.02) (1408,65.55) (1536,64.84) (1664,65.08) (1792,65.23) (1920,65.52) (2048,65.43) (2176,65.49) (2304,65.1) (2362,65.2) (2362,57.16) (2304,57.03) (2176,57.31) (2048,57.32) (1920,57.24) (1792,56.81) (1664,56.85) (1536,56.97) (1408,57.67) (1280,58.36) (1152,58.77) (1024,58.5) (896,58.04) (768,57.03) (640,57.19) (512,58.4) (384,58.59) (256,57.81) (224,58.04) (192,59.9) (160,58.75) (128,58.59) (96,56.25) (64,56.25) (32,46.88)} -- cycle;
\addplot[basecurve, line width=1.2pt] coordinates {(32,46.88) (64,56.25) (96,56.25) (128,58.59) (160,58.75) (192,59.9) (224,58.04) (256,57.81) (384,58.59) (512,58.4) (640,57.19) (768,57.03) (896,58.04) (1024,58.5) (1152,58.77) (1280,58.36) (1408,57.67) (1536,56.97) (1664,56.85) (1792,56.81) (1920,57.24) (2048,57.32) (2176,57.31) (2304,57.03) (2362,57.16)};
\addplot[oursframe, line width=1.7pt] coordinates {(32,53.12) (64,60.94) (96,61.46) (128,64.84) (160,65) (192,64.58) (224,64.73) (256,64.45) (384,64.58) (512,63.87) (640,64.22) (768,65.76) (896,66.63) (1024,66.6) (1152,66.15) (1280,66.02) (1408,65.55) (1536,64.84) (1664,65.08) (1792,65.23) (1920,65.52) (2048,65.43) (2176,65.49) (2304,65.1) (2362,65.2)};
\addplot[basecurve, only marks, mark=*, mark size=1.3pt, forget plot] coordinates {(2362,57.16)};
\addplot[oursframe, only marks, mark=*, mark size=1.5pt, forget plot] coordinates {(2362,65.2)};
\draw[axisgray, line width=0.5pt] (axis cs:0,44.5) -- (axis cs:2362,44.5);
\nextgroupplot[title={Task 2}, xmin=0, xmax=2503.7, ymin=0.173, ymax=0.388, xtick={0,1000,2000}, ytick={0.18,0.24,0.3,0.36}]
\addplot[fill=oursframe, fill opacity=0.13, draw=none, forget plot] coordinates {(85.89,0.215) (96,0.225) (128,0.283) (160,0.303) (192,0.318) (224,0.327) (256,0.327) (384,0.352) (512,0.364) (640,0.362) (768,0.363) (896,0.364) (1024,0.364) (1152,0.363) (1280,0.366) (1408,0.365) (1536,0.365) (1664,0.364) (1792,0.365) (1920,0.366) (2048,0.366) (2176,0.367) (2304,0.366) (2362,0.366) (2362,0.239) (2304,0.238) (2176,0.239) (2048,0.24) (1920,0.24) (1792,0.238) (1664,0.237) (1536,0.236) (1408,0.237) (1280,0.234) (1152,0.231) (1024,0.231) (896,0.231) (768,0.234) (640,0.234) (512,0.243) (384,0.236) (256,0.217) (224,0.22) (192,0.213) (160,0.213) (128,0.218) (96,0.214) (85.89,0.215)} -- cycle;
\addplot[basecurve, line width=1.2pt] coordinates {(32,0.224) (64,0.218) (96,0.214) (128,0.218) (160,0.213) (192,0.213) (224,0.22) (256,0.217) (384,0.236) (512,0.243) (640,0.234) (768,0.234) (896,0.231) (1024,0.231) (1152,0.231) (1280,0.234) (1408,0.237) (1536,0.236) (1664,0.237) (1792,0.238) (1920,0.24) (2048,0.24) (2176,0.239) (2304,0.238) (2362,0.239)};
\addplot[oursframe, line width=1.7pt] coordinates {(32,0.222) (64,0.193) (96,0.225) (128,0.283) (160,0.303) (192,0.318) (224,0.327) (256,0.327) (384,0.352) (512,0.364) (640,0.362) (768,0.363) (896,0.364) (1024,0.364) (1152,0.363) (1280,0.366) (1408,0.365) (1536,0.365) (1664,0.364) (1792,0.365) (1920,0.366) (2048,0.366) (2176,0.367) (2304,0.366) (2362,0.366)};
\addplot[basecurve, only marks, mark=*, mark size=1.3pt, forget plot] coordinates {(2362,0.239)};
\addplot[oursframe, only marks, mark=*, mark size=1.5pt, forget plot] coordinates {(2362,0.366)};
\draw[axisgray, line width=0.5pt] (axis cs:0,0.173) -- (axis cs:2362,0.173);
\nextgroupplot[title={Task 3}, xmin=0, xmax=734.6, ymin=0.498, ymax=0.664, xtick={0,300,600}, ytick={0.5,0.55,0.6,0.65}]
\addplot[fill=oursframe, fill opacity=0.13, draw=none, forget plot] coordinates {(32,0.583) (64,0.619) (96,0.633) (128,0.618) (160,0.616) (192,0.62) (224,0.607) (256,0.613) (384,0.635) (512,0.648) (640,0.645) (693,0.637) (693,0.514) (640,0.522) (512,0.528) (384,0.538) (256,0.532) (224,0.521) (192,0.529) (160,0.52) (128,0.534) (96,0.587) (64,0.596) (32,0.569)} -- cycle;
\addplot[basecurve, line width=1.2pt] coordinates {(32,0.569) (64,0.596) (96,0.587) (128,0.534) (160,0.52) (192,0.529) (224,0.521) (256,0.532) (384,0.538) (512,0.528) (640,0.522) (693,0.514)};
\addplot[oursframe, line width=1.7pt] coordinates {(32,0.583) (64,0.619) (96,0.633) (128,0.618) (160,0.616) (192,0.62) (224,0.607) (256,0.613) (384,0.635) (512,0.648) (640,0.645) (693,0.637)};
\addplot[basecurve, only marks, mark=*, mark size=1.3pt, forget plot] coordinates {(693,0.514)};
\addplot[oursframe, only marks, mark=*, mark size=1.5pt, forget plot] coordinates {(693,0.637)};
\draw[axisgray, line width=0.5pt] (axis cs:0,0.498) -- (axis cs:693,0.498);
\nextgroupplot[title={Task 4}, xmin=0, xmax=2503.7, ymin=0.336, ymax=0.438, xtick={0,1000,2000}, ytick={0.36,0.39,0.42}]
\addplot[fill=oursframe, fill opacity=0.13, draw=none, forget plot] coordinates {(32,0.394) (64,0.388) (96,0.396) (128,0.405) (160,0.411) (192,0.424) (224,0.428) (256,0.423) (384,0.422) (512,0.418) (640,0.417) (768,0.419) (896,0.409) (1024,0.413) (1152,0.408) (1280,0.408) (1408,0.409) (1536,0.41) (1664,0.41) (1792,0.413) (1920,0.413) (2048,0.413) (2176,0.413) (2304,0.413) (2362,0.414) (2362,0.386) (2304,0.386) (2176,0.384) (2048,0.383) (1920,0.382) (1792,0.38) (1664,0.379) (1536,0.379) (1408,0.379) (1280,0.379) (1152,0.379) (1024,0.382) (896,0.378) (768,0.373) (640,0.381) (512,0.397) (384,0.398) (256,0.393) (224,0.392) (192,0.383) (160,0.369) (128,0.358) (96,0.346) (64,0.35) (32,0.369)} -- cycle;
\addplot[basecurve, line width=1.2pt] coordinates {(32,0.369) (64,0.35) (96,0.346) (128,0.358) (160,0.369) (192,0.383) (224,0.392) (256,0.393) (384,0.398) (512,0.397) (640,0.381) (768,0.373) (896,0.378) (1024,0.382) (1152,0.379) (1280,0.379) (1408,0.379) (1536,0.379) (1664,0.379) (1792,0.38) (1920,0.382) (2048,0.383) (2176,0.384) (2304,0.386) (2362,0.386)};
\addplot[oursframe, line width=1.7pt] coordinates {(32,0.394) (64,0.388) (96,0.396) (128,0.405) (160,0.411) (192,0.424) (224,0.428) (256,0.423) (384,0.422) (512,0.418) (640,0.417) (768,0.419) (896,0.409) (1024,0.413) (1152,0.408) (1280,0.408) (1408,0.409) (1536,0.41) (1664,0.41) (1792,0.413) (1920,0.413) (2048,0.413) (2176,0.413) (2304,0.413) (2362,0.414)};
\addplot[basecurve, only marks, mark=*, mark size=1.3pt, forget plot] coordinates {(2362,0.386)};
\addplot[oursframe, only marks, mark=*, mark size=1.5pt, forget plot] coordinates {(2362,0.414)};
\draw[axisgray, line width=0.5pt] (axis cs:0,0.336) -- (axis cs:2362,0.336);
\nextgroupplot[title={Task 5}, xmin=0, xmax=2503.7, ymin=0.507, ymax=0.613, xtick={0,1000,2000}, ytick={0.51,0.54,0.57,0.6}]
\addplot[fill=oursframe, fill opacity=0.13, draw=none, forget plot] coordinates {(91.5,0.524) (96,0.525) (128,0.55) (160,0.561) (192,0.577) (224,0.589) (256,0.592) (384,0.597) (512,0.593) (640,0.586) (768,0.59) (896,0.595) (1024,0.592) (1152,0.594) (1280,0.592) (1408,0.59) (1536,0.588) (1664,0.592) (1792,0.601) (1920,0.599) (2048,0.6) (2176,0.602) (2304,0.603) (2362,0.602) (2362,0.548) (2304,0.549) (2176,0.546) (2048,0.543) (1920,0.541) (1792,0.544) (1664,0.542) (1536,0.544) (1408,0.541) (1280,0.539) (1152,0.539) (1024,0.534) (896,0.536) (768,0.536) (640,0.526) (512,0.536) (384,0.551) (256,0.545) (224,0.552) (192,0.544) (160,0.524) (128,0.517) (96,0.521) (91.5,0.524)} -- cycle;
\addplot[basecurve, line width=1.2pt] coordinates {(32,0.571) (64,0.543) (96,0.521) (128,0.517) (160,0.524) (192,0.544) (224,0.552) (256,0.545) (384,0.551) (512,0.536) (640,0.526) (768,0.536) (896,0.536) (1024,0.534) (1152,0.539) (1280,0.539) (1408,0.541) (1536,0.544) (1664,0.542) (1792,0.544) (1920,0.541) (2048,0.543) (2176,0.546) (2304,0.549) (2362,0.548)};
\addplot[oursframe, line width=1.7pt] coordinates {(32,0.557) (64,0.519) (96,0.525) (128,0.55) (160,0.561) (192,0.577) (224,0.589) (256,0.592) (384,0.597) (512,0.593) (640,0.586) (768,0.59) (896,0.595) (1024,0.592) (1152,0.594) (1280,0.592) (1408,0.59) (1536,0.588) (1664,0.592) (1792,0.601) (1920,0.599) (2048,0.6) (2176,0.602) (2304,0.603) (2362,0.602)};
\addplot[basecurve, only marks, mark=*, mark size=1.3pt, forget plot] coordinates {(2362,0.548)};
\addplot[oursframe, only marks, mark=*, mark size=1.5pt, forget plot] coordinates {(2362,0.602)};
\draw[axisgray, line width=0.5pt] (axis cs:0,0.507) -- (axis cs:2362,0.507);
\end{groupplot}
\end{tikzpicture}
\caption{Online scores of Base and Ours with GPT-5.6-Sol along the fixed stream. The x-axis counts processed cases, and each point is the score over all of them. The top row uses the metrics in Section~\ref{sec:experimental_setup}; the bottom row shows the five MedChain tasks, whose metrics are listed in Table~\ref{tab:medchain_online}. The last points equal the scores in Tables~\ref{tab:online_results} and~\ref{tab:medchain_online}. The shaded area marks where Ours is above Base.}
\label{fig:online_curves}
\end{figure}
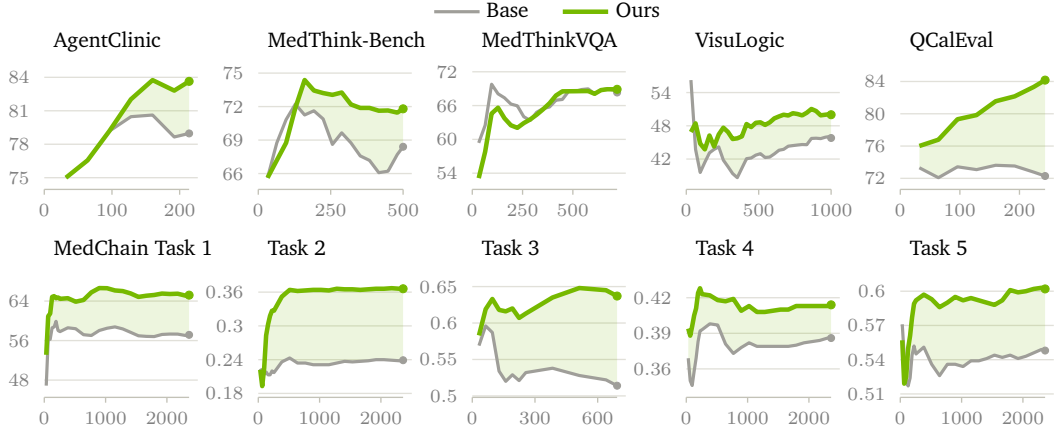

This comparison tests whether earlier experience improves later decisions. Each method starts with its own initial state and follows the same complete stream. Expertise is fixed within a batch, and the full-stream score includes the initial period before any update has been accepted. Figure~\ref{fig:online_curves} shows the scores of Base and Ours with GPT-5.6-Sol as more cases are processed.

\paragraph{Ours improves every base model on every benchmark.}
Ours scores above Base in all 24 benchmark--model pairs and achieves the best score in 22 of them (Table~\ref{tab:online_results}). Its average gain over Base is 17.6\%, compared with 5.6\% for ACE, which falls below Base in 7 pairs; where ACE scores higher, the margin is small (0.404 vs.\ 0.400 and 26.4 vs.\ 26.1). We attribute this stability to the validation strategy: each update is kept only if it helps on new cases without degrading performance on earlier ones, whereas ACE adds new experience without such a check.

\paragraph{Gains are largest where learned expertise can be reused.}
Gains reach 34.2\% on AgentClinic and 81.9\% on VisuLogic, and exceed 16\% on QCalEval for every base model. These benchmarks involve multi-step consultations or recurring question types, where a learned procedure or visual reference is likely to help many later cases. Gains are smaller on MedThink-Bench and MedThinkVQA (0.4\%--5.0\%), whose independent questions likely depend more on the model's own medical knowledge. Figure~\ref{fig:online_curves} shows that the gap between Ours and Base usually opens after the first few batches and then persists.
\paragraph{The improvement is robust to case order.}
Because the framework learns from a stream, its gains could depend on the order of cases. We therefore repeat the online experiment with GPT-5.6-Sol on AgentClinic and QCalEval under five random case orders (Table~\ref{tab:order_robustness}). Ours improves on Base by 5.7\% and 16.5\% on average, and both gaps are at least five times the largest standard deviation. The fixed-order results in Table~\ref{tab:online_results} also lie within one standard deviation of these means.

\begin{table*}[t]
\centering
\small
\setlength{\tabcolsep}{3pt}
\setlength{\aboverulesep}{0pt}
\setlength{\belowrulesep}{0pt}
\setlength{\arrayrulewidth}{1.2pt}
\renewcommand{\arraystretch}{1.15}

\begin{adjustbox}{scale=\restablescale}
\begin{tabular}{!{\vrule width 0.8pt}c ccO ccO ccO ccO}
\toprule
\rule{0pt}{17pt}\multirow{2}{*}{Benchmark}
&
\multicolumn{3}{c}{\begin{tabular}{@{}c@{}}Nemotron-3-Nano-\\Omni-30B\end{tabular}} &
\multicolumn{3}{c}{GPT-4o-mini} &
\multicolumn{3}{c}{Qwen3.8-27B} &
\multicolumn{3}{c!{\color{black}\vrule width 0.8pt}}{GPT-5.6-Sol} \\
\cmidrule(lr){2-4}
\cmidrule(lr){5-7}
\cmidrule(lr){8-10}
\cmidrule(lr){11-13}
\arrayrulecolor{oursframe}\cline{4-4}\cline{7-7}\cline{10-10}\cline{13-13}\arrayrulecolor{black}
& Base & ACE & Ours
& Base & ACE & Ours
& Base & ACE & Ours
& Base & ACE & Ours \\
\midrule

\rule{0pt}{2.6ex}\multirow{2}{*}{AgentClinic}
& 42.1 & 43.9 & \cellcolor{bestours}\textbf{56.5}
& 42.5 & 43.5 & \cellcolor{bestours}\textbf{54.2}
& 67.3 & 66.8 & \cellcolor{bestours}\textbf{73.4}
& 79.0 & 73.4 & \cellcolor{bestours}\textbf{83.6} \\
&  & \relup{4.3} & \cellcolor{bestours}\relup{34.2}
&  & \relup{2.4} & \cellcolor{bestours}\relup{27.5}
&  & \reldown{0.7} & \cellcolor{bestours}\relup{9.1}
&  & \reldown{7.1} & \cellcolor{bestours}\relup{5.8} \\[2pt]
\arrayrulecolor{rowsep}\specialrule{0.4pt}{0pt}{0pt}\arrayrulecolor{black}

\rule{0pt}{2.6ex}\multirow{2}{*}{MedChain}
& 0.334 & 0.341 & \cellcolor{bestours}\textbf{0.375}
& 0.353 & \cellcolor{bestother}\textbf{0.404} & 0.400
& 0.413 & 0.443 & \cellcolor{bestours}\textbf{0.460}
& 0.435 & 0.497 & \cellcolor{bestours}\textbf{0.513} \\
&  & \relup{2.1} & \cellcolor{bestours}\relup{12.3}
&  & \cellcolor{bestother}\relup{14.4} & \relup{13.3}
&  & \relup{7.3} & \cellcolor{bestours}\relup{11.4}
&  & \relup{14.3} & \cellcolor{bestours}\relup{17.9} \\[2pt]
\arrayrulecolor{rowsep}\specialrule{0.4pt}{0pt}{0pt}\arrayrulecolor{black}

\rule{0pt}{2.6ex}MedThink-
& 42.2 & 42.2 & \cellcolor{bestours}\textbf{42.8}
& 37.8 & 38.7 & \cellcolor{bestours}\textbf{39.0}
& 64.0 & 59.4 & \cellcolor{bestours}\textbf{64.6}
& 68.4 & 70.2 & \cellcolor{bestours}\textbf{71.8} \\
Bench &  & \relzero & \cellcolor{bestours}\relup{1.4}
&  & \relup{2.4} & \cellcolor{bestours}\relup{3.2}
&  & \reldown{7.2} & \cellcolor{bestours}\relup{0.9}
&  & \relup{2.6} & \cellcolor{bestours}\relup{5.0} \\[2pt]
\arrayrulecolor{rowsep}\specialrule{0.4pt}{0pt}{0pt}\arrayrulecolor{black}

\rule{0pt}{2.6ex}MedThink
& 25.4 & \cellcolor{bestother}\textbf{26.4} & 26.1
& 23.9 & 23.7 & \cellcolor{bestours}\textbf{24.5}
& 48.3 & 47.4 & \cellcolor{bestours}\textbf{48.5}
& 68.3 & 65.9 & \cellcolor{bestours}\textbf{68.9} \\
VQA &  & \cellcolor{bestother}\relup{3.9} & \relup{2.8}
&  & \reldown{0.8} & \cellcolor{bestours}\relup{2.5}
&  & \reldown{1.9} & \cellcolor{bestours}\relup{0.4}
&  & \reldown{3.5} & \cellcolor{bestours}\relup{0.9} \\[2pt]
\arrayrulecolor{rowsep}\specialrule{0.4pt}{0pt}{0pt}\arrayrulecolor{black}

\rule{0pt}{2.6ex}\multirow{2}{*}{VisuLogic}
& 14.4 & 21.4 & \cellcolor{bestours}\textbf{23.9}
& 26.6 & 23.6 & \cellcolor{bestours}\textbf{29.3}
& 21.5 & 29.9 & \cellcolor{bestours}\textbf{39.1}
& 45.8 & 47.3 & \cellcolor{bestours}\textbf{50.0} \\
&  & \relup{48.6} & \cellcolor{bestours}\relup{66.0}
&  & \reldown{11.3} & \cellcolor{bestours}\relup{10.2}
&  & \relup{39.1} & \cellcolor{bestours}\relup{81.9}
&  & \relup{3.3} & \cellcolor{bestours}\relup{9.2} \\[2pt]
\arrayrulecolor{rowsep}\specialrule{0.4pt}{0pt}{0pt}\arrayrulecolor{black}

\rule{0pt}{2.6ex}\multirow{2}{*}{QCalEval}
& 47.3 & 48.6 & \cellcolor{bestours}\textbf{55.5}
& 40.9 & 46.0 & \cellcolor{bestours}\textbf{58.5}
& 62.9 & 63.4 & \cellcolor{bestours}\textbf{82.3}
& 72.3 & 76.9 & \cellcolor{bestours}\textbf{84.2} \\
&  & \relup{2.7} & \cellcolor{bestours}\relup{17.3}
&  & \relup{12.5} & \cellcolor{bestours}\relup{43.0}
&  & \relup{0.8} & \cellcolor{bestours}\relup{30.8}
&  & \relup{6.4} & \cellcolor{bestours}\relup{16.5} \\[2pt]
\arrayrulecolor{oursframe}\cline{4-4}\cline{7-7}\cline{10-10}\cline{13-13}\arrayrulecolor{black}
\bottomrule
\end{tabular}
\end{adjustbox}
\caption{
Online performance on the fixed stream. Higher is better for all metrics. The number below each score gives its relative change from Base (green: gain; red: decline). The best score for each base model is shaded.
}
\label{tab:online_results}

\end{table*}

\subsection{Ablation Study}
\label{sec:component_ablations}

To isolate the role of each component, we compare the full framework with variants that remove Skill, Knowledge Memory, or MMKB. Each variant learns independently from the start of the same stream using one selected base model; it is not formed by removing a component after training. Removing Memory disables stored knowledge but keeps external knowledge searches, while removing MMKB disables visual retrieval and growth. Text-only tasks need no separate MMKB ablation. We also test a variant that accepts every update without validation (Section~\ref{sec:joint_validation}). We use GPT-5.6-Sol as the base model, and Table~\ref{tab:e2} reports the results.

\paragraph{Each component accounts for a different part of the gain.}
Among the three components, removing the Skill causes the largest drops on text-based and multi-step tasks, 12.1\% on MedChain and 5.5\% on AgentClinic, where the score returns to Base (Table~\ref{tab:e2}). Removing MMKB causes the largest drops on visual tasks, 14.0\% on QCalEval and 8.2\% on VisuLogic, and brings both back to their Base scores (72.4 vs.\ 72.3 and 45.9 vs.\ 45.8), so nearly all of the visual gain comes from MMKB. Removing Knowledge Memory has a smaller effect, largest on MedThink-Bench (2.5\%), which tests medical knowledge. Within MedChain, removing MMKB changes only Task~3, the task with images, whereas removing the Skill lowers all five tasks (Table~\ref{tab:ablation_details}). The three components are therefore complementary rather than redundant.

\paragraph{Validation is necessary for the gains.}
Accepting every update without validation lowers the score on all six benchmarks by 21.8\% to 37.8\% relative to Full and leaves every score below Base (Table~\ref{tab:e2}). Updates that hurt performance are then kept and used on later cases, so validation is needed for the learned expertise to help rather than harm.

\paragraph{MMKB versus standard multimodal retrieval.}
We also test whether MMKB's reference analysis adds value beyond standard multimodal retrieval. On image-based tasks, both conditions use GPT-5.6-Sol under the online protocol, keep all other components and task settings the same, and receive identical retrieved references with images, context, and source answers. The baseline uses these references as context and can still reason before answering, but it is not asked to analyze each reference or compare it with the current case. We also test Initial MMKB, which uses only the initial references (Table~\ref{tab:mmkb_sources}) and does not add deployment cases. Figure~\ref{fig:mmkb_retrieval} reports the results.

\paragraph{MMKB gains come from how references are used.}
With identical references, standard retrieval changes the score by $-3.5\%$ to $+5.0\%$ relative to Base, whereas MMKB improves all four image-based tasks, by up to 23.9\% on MedChain Task~3 (Figure~\ref{fig:mmkb_retrieval}). Standard retrieval even lowers the score on MedChain Task~3 and MedThinkVQA, suggesting that references given only as context can distract the model when their relation to the current image is not analyzed. MMKB turns the same references into gains by guiding the model to examine each reference and compare it with the current case. Initial MMKB still improves all four tasks, by up to 22.6\% on MedChain Task~3, and adding deployment cases increases the gain further on every task.

\begin{table}[t]
\centering
\begin{minipage}[t]{0.46\linewidth}
\vspace{0pt}
\centering
\begingroup
\small
\setlength{\tabcolsep}{6pt}
\setlength{\aboverulesep}{0pt}
\setlength{\belowrulesep}{0pt}
\setlength{\arrayrulewidth}{1.2pt}
\renewcommand{\arraystretch}{1.15}
\begin{adjustbox}{scale=\restablescale}
\begin{tabular}{!{\vrule width 0.8pt}ccO}
\toprule
\arrayrulecolor{oursframe}\cline{3-3}\arrayrulecolor{black}
\rule{0pt}{2.6ex}Benchmark & Base & Ours \\
\midrule
\rule{0pt}{2.6ex}AgentClinic & 79.3 $\pm$ 0.7 & \cellcolor{bestours}\textbf{83.8 $\pm$ 0.9} \\[2pt]
\arrayrulecolor{rowsep}\specialrule{0.4pt}{0pt}{0pt}\arrayrulecolor{black}
\rule{0pt}{2.6ex}QCalEval & 71.3 $\pm$ 1.8 & \cellcolor{bestours}\textbf{83.1 $\pm$ 1.5} \\[2pt]
\arrayrulecolor{oursframe}\cline{3-3}\arrayrulecolor{black}
\bottomrule
\end{tabular}
\end{adjustbox}
\caption{Online results of GPT-5.6-Sol under five random case orders (mean $\pm$ standard deviation).}
\label{tab:order_robustness}
\endgroup

\vspace{5pt}
\begin{tikzpicture}
\begin{axis}[
  scale only axis, width=0.71\linewidth, height=1.75cm,
  ybar=1.5pt, bar width=6pt,
  ymin=-6, ymax=44, ytick={0,20,40},
  symbolic x coords={MedChain,MedThinkVQA,VisuLogic,QCalEval},
  xtick=data,
  xticklabels={{MedChain\\Task 3},{MedThink\\VQA},{VisuLogic},{QCalEval}},
  xticklabel style={align=center, font=\scriptsize, text=black!75},
  yticklabel style={font=\footnotesize, text=black!50},
  ylabel={Improvement over\\Base (\%)}, ylabel style={font=\footnotesize, text=black!70, align=center},
  enlarge x limits=0.08,
  axis x line*=bottom, axis y line*=left,
  x axis line style={draw=none}, y axis line style={draw=none},
  tick style={draw=none},
  ymajorgrids, grid style={draw=rowsep, line width=0.5pt},
  extra y ticks={0}, extra y tick labels={}, extra y tick style={grid=major, major grid style={draw=black!45, line width=0.9pt}},
  clip mode=individual,
  point meta=y,
  nodes near coords={\pgfmathprintnumber[fixed, fixed zerofill, precision=1, showpos]{\pgfplotspointmeta}\%},
  every node near coord/.append style={font=\fontsize{5}{6}\selectfont, text=black!75, rotate=90, anchor=west, inner sep=1pt},
  legend style={at={(0.38,1.03)}, anchor=south, draw=none, legend columns=3, font=\scriptsize, /tikz/every even column/.append style={column sep=5pt}},
  legend image code/.code={\fill[#1] (0cm,-0.08cm) rectangle (0.22cm,0.14cm);},
]
\addplot[fill=stdretr, draw=none] coordinates {(MedChain,-3.502) (MedThinkVQA,-2.196) (VisuLogic,0.655) (QCalEval,4.979)};
\addplot[fill=ourslight, draw=none] coordinates {(MedChain,22.568) (MedThinkVQA,0.293) (VisuLogic,8.515) (QCalEval,15.214)};
\addplot[fill=oursframe, draw=none] coordinates {(MedChain,23.930) (MedThinkVQA,0.878) (VisuLogic,9.170) (QCalEval,16.459)};
\legend{Standard retrieval, Initial MMKB, MMKB}
\end{axis}
\end{tikzpicture}%

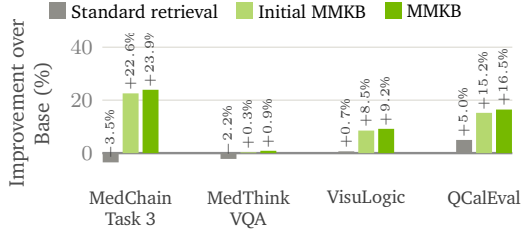
\captionof{figure}{Standard multimodal retrieval versus MMKB with GPT-5.6-Sol. Initial MMKB does not add deployment cases. Absolute scores are in Table~\ref{tab:mmkb_retrieval_scores}.}
\label{fig:mmkb_retrieval}
\end{minipage}\hfill
\begin{minipage}[t]{0.51\linewidth}
\vspace{0pt}
\centering
\small
\setlength{\tabcolsep}{3pt}
\setlength{\aboverulesep}{0pt}
\setlength{\belowrulesep}{0pt}
\setlength{\arrayrulewidth}{1.2pt}
\renewcommand{\arraystretch}{1.15}
\begin{adjustbox}{scale=\restablescale}
\begin{tabular}{!{\vrule width 0.8pt}ccOcccc!{\vrule width 0.8pt}}
\toprule
\arrayrulecolor{oursframe}\cline{3-3}\arrayrulecolor{black}
\rule{0pt}{2.6ex}Benchmark & Base & Full & $-$Skill & $-$Memory & $-$MMKB & $-$Validation \\
\midrule
\rule{0pt}{2.6ex}\multirow{2}{*}{AgentClinic} & 79.0 & 83.6 & 79.0 & 83.2 & = Full & 53.5 \\
 &  &  & \reldown{5.5} & \reldown{0.5} &  & \reldown{36.0} \\[2pt]
\arrayrulecolor{rowsep}\specialrule{0.4pt}{0pt}{0pt}\arrayrulecolor{black}

\rule{0pt}{2.6ex}\multirow{2}{*}{MedChain} & 0.435 & 0.513 & 0.451 & 0.505 & 0.494 & 0.401 \\
 &  &  & \reldown{12.1} & \reldown{1.6} & \reldown{3.7} & \reldown{21.8} \\[2pt]
\arrayrulecolor{rowsep}\specialrule{0.4pt}{0pt}{0pt}\arrayrulecolor{black}

\rule{0pt}{2.6ex}MedThink- & 68.4 & 71.8 & 69.4 & 70.0 & = Full & 51.2 \\
Bench &  &  & \reldown{3.3} & \reldown{2.5} &  & \reldown{28.7} \\[2pt]
\arrayrulecolor{rowsep}\specialrule{0.4pt}{0pt}{0pt}\arrayrulecolor{black}

\rule{0pt}{2.6ex}MedThink & 68.3 & 68.9 & 68.9 & 68.9 & 68.3 & 52.1 \\
VQA &  &  & \relzero & \relzero & \reldown{0.9} & \reldown{24.4} \\[2pt]
\arrayrulecolor{rowsep}\specialrule{0.4pt}{0pt}{0pt}\arrayrulecolor{black}

\rule{0pt}{2.6ex}\multirow{2}{*}{VisuLogic} & 45.8 & 50.0 & 49.6 & 50.0 & 45.9 & 38.1 \\
 &  &  & \reldown{0.8} & \relzero & \reldown{8.2} & \reldown{23.8} \\[2pt]
\arrayrulecolor{rowsep}\specialrule{0.4pt}{0pt}{0pt}\arrayrulecolor{black}

\rule{0pt}{2.6ex}\multirow{2}{*}{QCalEval} & 72.3 & 84.2 & 82.8 & 84.2 & 72.4 & 52.4 \\
 &  &  & \reldown{1.7} & \relzero & \reldown{14.0} & \reldown{37.8} \\[2pt]
\arrayrulecolor{oursframe}\cline{3-3}\arrayrulecolor{black}
\bottomrule
\end{tabular}
\end{adjustbox}
\caption{Online component ablation with GPT-5.6-Sol. Full is the complete framework; $-$Skill, $-$Memory, and $-$MMKB remove the Skill, the stored Knowledge Memory (external searches remain), and MMKB, respectively; $-$Validation accepts every update without validation. The number below each ablated score gives its relative change from Full. $-$MMKB equals Full on tasks that do not use MMKB.}
\label{tab:e2}
\end{minipage}
\end{table}

\subsection{Frozen Generalization to Unseen Cases}
\label{sec:frozen_generalization}


To separate generalization from continued adaptation, each method learns from approximately the first 70\% of the fixed stream and is then evaluated on the remaining cases with learning disabled. Cases sharing a patient, source, or experiment identity stay in the same split. We freeze the final Skill, Knowledge Memory, and MMKB, including only visual additions from the learning prefix. Test feedback is not available to the learner. 

\paragraph{The learned expertise generalizes to unseen cases.}
After being frozen, Ours scores above Base on the test split in 21 of 24 pairs and achieves the best score in 19 (Table~\ref{tab:frozen_results}). Its average gain over Base is 20.7\%, compared with 8.3\% for SkillOpt and 1.9\% for ACE. This gain is comparable to the online gain in Section~\ref{sec:online_improvement}, suggesting that the expertise captures reusable procedures and knowledge rather than corrections to the learning cases. The baselines are less stable on unseen cases: SkillOpt falls below Base in 7 pairs and ACE in 12, whereas Ours falls below Base only once ($-3.6\%$ on MedThinkVQA with Nemotron-3-Nano-Omni-30B). This matches the motivation of our validation strategy. SkillOpt repeatedly selects updates on a fixed validation set, and ACE adds experience without validation, so their updates can fit the learning cases without carrying over to new ones.

\begin{table*}[t]
\centering
\small
\setlength{\tabcolsep}{3pt}
\setlength{\aboverulesep}{0pt}
\setlength{\belowrulesep}{0pt}
\setlength{\arrayrulewidth}{1.2pt}
\renewcommand{\arraystretch}{1.15}

\begin{adjustbox}{scale=\restablescale}
\begin{tabular}{!{\vrule width 0.8pt}c cccO cccO cccO cccO}
\toprule
\rule{0pt}{17pt}\multirow{2}{*}{Benchmark}
&
\multicolumn{4}{c}{\begin{tabular}{@{}c@{}}Nemotron-3-Nano-\\Omni-30B\end{tabular}} &
\multicolumn{4}{c}{GPT-4o-mini} &
\multicolumn{4}{c}{Qwen3.8-27B} &
\multicolumn{4}{c!{\color{black}\vrule width 0.8pt}}{GPT-5.6-Sol} \\
\cmidrule(lr){2-5}
\cmidrule(lr){6-9}
\cmidrule(lr){10-13}
\cmidrule(lr){14-17}
\arrayrulecolor{oursframe}\cline{5-5}\cline{9-9}\cline{13-13}\cline{17-17}\arrayrulecolor{black}
& Base & SkillOpt & ACE & Ours
& Base & SkillOpt & ACE & Ours
& Base & SkillOpt & ACE & Ours
& Base & SkillOpt & ACE & Ours \\
\midrule

\rule{0pt}{2.6ex}\multirow{2}{*}{AgentClinic}
& 40.0 & 39.4 & 40.2 & \cellcolor{bestours}\textbf{58.5}
& 38.5 & \cellcolor{bestother}\textbf{53.8} & 41.5 & 52.3
& 67.7 & 67.7 & 64.6 & \cellcolor{bestours}\textbf{73.8}
& 76.9 & 75.4 & 72.3 & \cellcolor{bestours}\textbf{80.0} \\
&  & \reldown{1.5} & \relup{0.5} & \cellcolor{bestours}\relup{46.3}
&  & \cellcolor{bestother}\relup{39.7} & \relup{7.8} & \relup{35.8}
&  & \relzero & \reldown{4.6} & \cellcolor{bestours}\relup{9.0}
&  & \reldown{2.0} & \reldown{6.0} & \cellcolor{bestours}\relup{4.0} \\[2pt]
\arrayrulecolor{rowsep}\specialrule{0.4pt}{0pt}{0pt}\arrayrulecolor{black}

\rule{0pt}{2.6ex}\multirow{2}{*}{MedChain}
& 0.336 & 0.367 & 0.331 & \cellcolor{bestours}\textbf{0.377}
& 0.356 & 0.379 & 0.330 & \cellcolor{bestours}\textbf{0.389}
& 0.431 & 0.450 & 0.433 & \cellcolor{bestours}\textbf{0.456}
& 0.442 & \cellcolor{bestother}\textbf{0.526} & 0.497 & 0.520 \\
&  & \relup{9.2} & \reldown{1.5} & \cellcolor{bestours}\relup{12.2}
&  & \relup{6.5} & \reldown{7.3} & \cellcolor{bestours}\relup{9.3}
&  & \relup{4.4} & \relup{0.5} & \cellcolor{bestours}\relup{5.8}
&  & \cellcolor{bestother}\relup{19.0} & \relup{12.4} & \relup{17.6} \\[2pt]
\arrayrulecolor{rowsep}\specialrule{0.4pt}{0pt}{0pt}\arrayrulecolor{black}

\rule{0pt}{2.6ex}MedThink-
& 42.7 & \cellcolor{bestother}\textbf{44.7} & 43.7 & 42.7
& 34.7 & 36.7 & 38.0 & \cellcolor{bestours}\textbf{39.3}
& 64.7 & 57.7 & 60.4 & \cellcolor{bestours}\textbf{66.0}
& 72.0 & 72.4 & 71.2 & \cellcolor{bestours}\textbf{76.0} \\
Bench &  & \cellcolor{bestother}\relup{4.7} & \relup{2.3} & \relzero
&  & \relup{5.8} & \relup{9.5} & \cellcolor{bestours}\relup{13.3}
&  & \reldown{10.8} & \reldown{6.6} & \cellcolor{bestours}\relup{2.0}
&  & \relup{0.6} & \reldown{1.1} & \cellcolor{bestours}\relup{5.6} \\[2pt]
\arrayrulecolor{rowsep}\specialrule{0.4pt}{0pt}{0pt}\arrayrulecolor{black}

\rule{0pt}{2.6ex}MedThink
& 25.0 & \cellcolor{bestother}\textbf{25.0} & 21.3 & 24.1
& 23.0 & 21.9 & 21.3 & \cellcolor{bestours}\textbf{23.0}
& 51.4 & \cellcolor{bestother}\textbf{53.7} & 51.6 & 53.2
& 69.0 & 69.4 & 67.7 & \cellcolor{bestours}\textbf{69.4} \\
VQA &  & \cellcolor{bestother}\relzero & \reldown{14.8} & \reldown{3.6}
&  & \reldown{4.8} & \reldown{7.4} & \cellcolor{bestours}\relzero
&  & \cellcolor{bestother}\relup{4.5} & \relup{0.4} & \relup{3.5}
&  & \relup{0.6} & \reldown{1.9} & \cellcolor{bestours}\relup{0.6} \\[2pt]
\arrayrulecolor{rowsep}\specialrule{0.4pt}{0pt}{0pt}\arrayrulecolor{black}

\rule{0pt}{2.6ex}\multirow{2}{*}{VisuLogic}
& 14.3 & 23.1 & 15.3 & \cellcolor{bestours}\textbf{23.7}
& 27.0 & 22.7 & 21.7 & \cellcolor{bestours}\textbf{27.3}
& 21.7 & 27.8 & 30.2 & \cellcolor{bestours}\textbf{41.7}
& 49.3 & 48.0 & 46.3 & \cellcolor{bestours}\textbf{50.7} \\
&  & \relup{61.5} & \relup{7.0} & \cellcolor{bestours}\relup{65.7}
&  & \reldown{15.9} & \reldown{19.6} & \cellcolor{bestours}\relup{1.1}
&  & \relup{28.1} & \relup{39.2} & \cellcolor{bestours}\relup{92.2}
&  & \reldown{2.6} & \reldown{6.1} & \cellcolor{bestours}\relup{2.8} \\[2pt]
\arrayrulecolor{rowsep}\specialrule{0.4pt}{0pt}{0pt}\arrayrulecolor{black}

\rule{0pt}{2.6ex}\multirow{2}{*}{QCalEval}
& 43.2 & 42.7 & 41.0 & \cellcolor{bestours}\textbf{54.8}
& 36.8 & 53.4 & 52.1 & \cellcolor{bestours}\textbf{61.1}
& 58.8 & 62.4 & 59.2 & \cellcolor{bestours}\textbf{87.9}
& 69.0 & 70.8 & 72.2 & \cellcolor{bestours}\textbf{89.7} \\
&  & \reldown{1.2} & \reldown{5.1} & \cellcolor{bestours}\relup{26.9}
&  & \relup{45.1} & \relup{41.6} & \cellcolor{bestours}\relup{66.0}
&  & \relup{6.1} & \relup{0.7} & \cellcolor{bestours}\relup{49.5}
&  & \relup{2.6} & \relup{4.6} & \cellcolor{bestours}\relup{30.0} \\[2pt]
\arrayrulecolor{oursframe}\cline{5-5}\cline{9-9}\cline{13-13}\cline{17-17}\arrayrulecolor{black}
\bottomrule
\end{tabular}
\end{adjustbox}
\caption{
Frozen generalization after learning on the first approximately 70\% of the fixed stream and evaluating on the remaining cases. Source groups are kept intact. Higher is better for all metrics. The number below each score gives its relative change from Base (green: gain; red: decline). The best score for each base model is shaded.
}
\label{tab:frozen_results}

\end{table*}

\subsection{Zero-Step Cross-Model Transfer}
\label{sec:cross_model_transfer}


This experiment tests whether expertise learned with one model can improve a different model without further learning. We use two source models. When GPT-5.6-Sol is the source, it serves as both the base model answering cases and the optimizer; when Qwen3.8-27B is the source, it answers cases and GPT-5.6-Sol serves as the optimizer. For each source, we freeze the final Skill, Knowledge Memory, and MMKB from its complete online run, then transfer them unchanged to the other three models.

\begin{figure}[t]
\centering
\begin{tikzpicture}
\begin{groupplot}[
  group style={group size=2 by 1, horizontal sep=0.35cm},
  set layers,
  scale only axis, width=0.435\linewidth, height=2.0cm,
  ybar=2pt, /pgf/bar width=6pt,
  ymin=-6, ymax=125, ytick={0,20,40,60},
  symbolic x coords={AC,MC,MTB,MTVQA,VL,QC},
  xtick=data,
  xticklabels={{AgentClinic},{MedChain},{MedThink-\\Bench},{MedThink\\VQA},{VisuLogic},{QCalEval}},
  xticklabel style={align=center, font=\fontsize{6}{7}\selectfont, text=black!75},
  yticklabel style={font=\footnotesize, text=black!50},
  ylabel={Improvement over\\Base (\%)}, ylabel style={font=\footnotesize, text=black!70, align=center},
  enlarge x limits=0.1,
  axis x line*=bottom, axis y line*=left,
  x axis line style={draw=none}, y axis line style={draw=none},
  tick style={draw=none},
  ymajorgrids, grid style={draw=rowsep, line width=0.5pt},
  extra y ticks={0}, extra y tick labels={}, extra y tick style={grid=major, major grid style={draw=black!45, line width=0.9pt}},
  clip mode=individual,
  point meta=y,
  nodes near coords={\pgfmathfloattofixed{\pgfplotspointmeta}\let\pmfix\pgfmathresult\pgfmathtruncatemacro\iszero{abs(\pmfix)<0.05}\ifnum\iszero=1 0.0\%\else\pgfmathprintnumber[fixed, fixed zerofill, precision=1, showpos]{\pgfplotspointmeta}\%\fi},
  every node near coord/.append style={font=\fontsize{5}{6}\selectfont, text=black!75, rotate=90, anchor=west, inner sep=1pt},
  legend style={at={(1.029,1.04)}, anchor=south, draw=none, legend columns=4, font=\footnotesize, /tikz/every even column/.append style={column sep=8pt}},
  legend image code/.code={\fill[#1] (0cm,-0.08cm) rectangle (0.22cm,0.14cm);},
]
\nextgroupplot
\begin{pgfonlayer}{axis background}
\fill[bandblue] (rel axis cs:0.1667,0) rectangle (rel axis cs:0.3333,1);
\fill[bandblue] (rel axis cs:0.5,0) rectangle (rel axis cs:0.6667,1);
\fill[bandblue] (rel axis cs:0.8333,0) rectangle (rel axis cs:1,1);
\end{pgfonlayer}
\addplot[fill=ourstext, draw=none] coordinates {(AC,2.1) (MC,10.2) (MTB,6.2) (MTVQA,5.9) (VL,68.8) (QC,33.6)};
\addplot[fill=oursframe, draw=none] coordinates {(AC,2.4) (MC,9.3) (MTB,3.7) (MTVQA,17.6) (VL,-1.1) (QC,66.7)};
\addplot[fill=ourslight, draw=none] coordinates {(AC,10.4) (MC,18.9) (MTB,5.0) (MTVQA,5.2) (VL,66.5) (QC,42.8)};
\addlegendimage{fill=ourstext!60!black, draw=none}
\legend{Nemotron-3-Nano-Omni-30B, GPT-4o-mini, Qwen3.8-27B, GPT-5.6-Sol}
\node[anchor=north west, inner sep=1pt, font=\footnotesize, text=black] at (rel axis cs:0,1) {(a)};
\nextgroupplot[ylabel={}, yticklabels={}]
\begin{pgfonlayer}{axis background}
\fill[bandblue] (rel axis cs:0.1667,0) rectangle (rel axis cs:0.3333,1);
\fill[bandblue] (rel axis cs:0.5,0) rectangle (rel axis cs:0.6667,1);
\fill[bandblue] (rel axis cs:0.8333,0) rectangle (rel axis cs:1,1);
\end{pgfonlayer}
\addplot[fill=ourstext, draw=none] coordinates {(AC,25.4) (MC,8.4) (MTB,12.8) (MTVQA,0.0) (VL,62.5) (QC,29.0)};
\addplot[fill=oursframe, draw=none] coordinates {(AC,47.3) (MC,9.1) (MTB,8.5) (MTVQA,7.1) (VL,2.3) (QC,65.8)};
\addplot[fill=ourstext!60!black, draw=none] coordinates {(AC,11.8) (MC,15.9) (MTB,12.9) (MTVQA,-0.7) (VL,6.3) (QC,23.2)};
\node[anchor=north west, inner sep=1pt, font=\footnotesize, text=black] at (rel axis cs:0,1) {(b)};
\end{groupplot}
\end{tikzpicture}%
\caption{Zero-step transfer with (a) GPT-5.6-Sol and (b) Qwen3.8-27B as the source. Bars show the relative change from each target's Base (the zero line); absolute scores are in Tables~\ref{tab:transfer_sol} and~\ref{tab:transfer_qwen}.}
\label{fig:cross_model_transfer}
\end{figure}
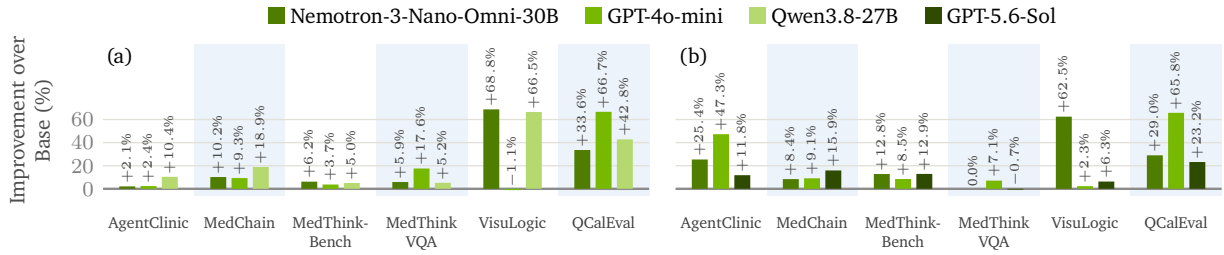

Each target is evaluated on the complete benchmark without further learning. References matching the current case or detected as duplicates remain excluded. We compare against the same target model's full-stream Base result under matching evaluation conditions. Figure~\ref{fig:cross_model_transfer} reports the results for both source models.

\paragraph{Expertise transfers across models without further learning.}
Transferred expertise improves the target model in 33 of the 36 transfer results, with an average gain of 20.0\% over the target's Base (Figure~\ref{fig:cross_model_transfer}). Expertise learned with the open-weight Qwen3.8-27B also improves the closed-source GPT-5.6-Sol, by 11.8\% on AgentClinic and 23.2\% on QCalEval, and gives the weaker GPT-4o-mini a 47.3\% gain on AgentClinic. Because the expertise is kept outside the model, a new or upgraded base model can use it immediately instead of learning it again.

\section{Conclusion}
\label{sec:conclusion}

We presented a framework that improves frozen LLMs and VLMs by accumulating external expertise during deployment. It maintains a Skill for reasoning and tool use, a Knowledge Memory for reliable facts, and an MMKB for visual cases with known answers, and it tests updates to the Skill and Knowledge Memory on new and past cases before deployment. Across six benchmarks, the framework improved performance during online deployment, generalized to unseen cases, transferred to other models without further optimization, and also worked in non-medical domains.

\bibliographystyle{plainnat}
\bibliography{references}

\clearpage
\appendix
\raggedbottom
\section{Detailed Experimental Results}
\label{sec:appendix_results}

\begin{table}[!h]
\centering
\small
\setlength{\tabcolsep}{6pt}
\setlength{\aboverulesep}{0pt}
\setlength{\belowrulesep}{0pt}
\setlength{\arrayrulewidth}{1.2pt}
\renewcommand{\arraystretch}{1.15}
\begin{adjustbox}{scale=\restablescale}
\begin{tabular}{!{\vrule width 0.8pt}c cc cc cc cc!{\vrule width 0.8pt}}
\toprule
\rule{0pt}{17pt}\multirow{2}{*}{MedChain}
&
\multicolumn{2}{c}{\begin{tabular}{@{}c@{}}Nemotron-3-Nano-\\Omni-30B\end{tabular}} &
\multicolumn{2}{c}{GPT-4o-mini} &
\multicolumn{2}{c}{Qwen3.8-27B} &
\multicolumn{2}{c!{\vrule width 0.8pt}}{GPT-5.6-Sol} \\
\cmidrule(lr){2-3}
\cmidrule(lr){4-5}
\cmidrule(lr){6-7}
\cmidrule(lr){8-9}
& Base & Ours
& Base & Ours
& Base & Ours
& Base & Ours \\
\midrule
\rule{0pt}{2.6ex}Task 1 & 53.8 / 0.243 & 58.0 / 0.305 & 59.1 / 0.304 & 59.9 / 0.366 & 58.9 / 0.360 & 63.0 / 0.393 & 57.2 / 0.351 & 65.2 / 0.406 \\[2pt]
\arrayrulecolor{rowsep}\specialrule{0.4pt}{0pt}{0pt}\arrayrulecolor{black}
\rule{0pt}{2.6ex}Task 2 & 0.225 & 0.289 & 0.174 & 0.323 & 0.241 & 0.340 & 0.239 & 0.366 \\[2pt]
\arrayrulecolor{rowsep}\specialrule{0.4pt}{0pt}{0pt}\arrayrulecolor{black}
\rule{0pt}{2.6ex}Task 3 & 0.238 & 0.244 & 0.190 & 0.203 & 0.410 & 0.416 & 0.514 & 0.637 \\[2pt]
\arrayrulecolor{rowsep}\specialrule{0.4pt}{0pt}{0pt}\arrayrulecolor{black}
\rule{0pt}{2.6ex}Task 4 & 0.353 & 0.360 & 0.362 & 0.363 & 0.374 & 0.414 & 0.386 & 0.414 \\[2pt]
\arrayrulecolor{rowsep}\specialrule{0.4pt}{0pt}{0pt}\arrayrulecolor{black}
\rule{0pt}{2.6ex}Task 5 & 0.406 & 0.473 & 0.500 & 0.543 & 0.501 & 0.564 & 0.548 & 0.602 \\[2pt]
\bottomrule
\end{tabular}
\end{adjustbox}
\caption{Per-task MedChain results during online deployment (Section~\ref{sec:online_improvement}). Task~1 reports first-level referral accuracy (\%) / second-level referral IoU, Task~2 history-taking examination-item IoU, Task~3 macro claim recall, Task~4 normalized diagnosis score, and Task~5 treatment IoU. The mean of these six metrics on a 0--1 scale gives the MedChain Average in Table~\ref{tab:online_results}.}
\label{tab:medchain_online}
\end{table}

\begin{table}[!h]
\centering
\small
\setlength{\tabcolsep}{6pt}
\setlength{\aboverulesep}{0pt}
\setlength{\belowrulesep}{0pt}
\setlength{\arrayrulewidth}{1.2pt}
\renewcommand{\arraystretch}{1.15}
\begin{adjustbox}{scale=\restablescale}
\begin{tabular}{!{\vrule width 0.8pt}ccOcccc!{\vrule width 0.8pt}}
\toprule
\arrayrulecolor{oursframe}\cline{3-3}\arrayrulecolor{black}
\rule{0pt}{2.6ex}MedChain & Base & Full & $-$Skill & $-$Memory & $-$MMKB & $-$Validation \\
\midrule
\rule{0pt}{2.6ex}Task 1 & 57.2 / 0.351 & 65.2 / 0.406 & 57.3 / 0.361 & 64.9 / 0.402 & = Full & 54.4 / 0.330 \\[2pt]
\arrayrulecolor{rowsep}\specialrule{0.4pt}{0pt}{0pt}\arrayrulecolor{black}
\rule{0pt}{2.6ex}Task 2 & 0.239 & 0.366 & 0.241 & 0.364 & = Full & 0.218 \\[2pt]
\arrayrulecolor{rowsep}\specialrule{0.4pt}{0pt}{0pt}\arrayrulecolor{black}
\rule{0pt}{2.6ex}Task 3 & 0.514 & 0.637 & 0.604 & 0.613 & 0.523 & 0.488 \\[2pt]
\arrayrulecolor{rowsep}\specialrule{0.4pt}{0pt}{0pt}\arrayrulecolor{black}
\rule{0pt}{2.6ex}Task 4 & 0.386 & 0.414 & 0.387 & 0.397 & = Full & 0.327 \\[2pt]
\arrayrulecolor{rowsep}\specialrule{0.4pt}{0pt}{0pt}\arrayrulecolor{black}
\rule{0pt}{2.6ex}Task 5 & 0.548 & 0.602 & 0.539 & 0.604 & = Full & 0.501 \\[2pt]
\arrayrulecolor{rowsep}\specialrule{0.4pt}{0pt}{0pt}\arrayrulecolor{black}
\rule{0pt}{2.6ex}Average & 0.435 & 0.513 & 0.451 & 0.505 & 0.494 & 0.401 \\[2pt]
\arrayrulecolor{oursframe}\cline{3-3}\arrayrulecolor{black}
\bottomrule
\end{tabular}
\end{adjustbox}
\caption{Per-task MedChain results of the online component ablation with GPT-5.6-Sol; the other benchmarks are reported in Table~\ref{tab:e2}. Task~1 reports first-level referral accuracy (\%) / second-level referral IoU, Task~2 history-taking examination-item IoU, Task~3 macro claim recall, Task~4 normalized diagnosis score, and Task~5 treatment IoU. Average is the mean of these six metrics on a 0--1 scale (Appendix~\ref{sec:appendix_method}). $-$MMKB equals Full on tasks that do not use MMKB.}
\label{tab:ablation_details}
\end{table}

\begin{table}[!h]
\centering
\small
\setlength{\tabcolsep}{6pt}
\setlength{\aboverulesep}{0pt}
\setlength{\belowrulesep}{0pt}
\setlength{\arrayrulewidth}{1.2pt}
\renewcommand{\arraystretch}{1.15}
\begin{adjustbox}{scale=\restablescale}
\begin{tabular}{!{\vrule width 0.8pt}ccccc!{\vrule width 0.8pt}}
\toprule
\rule{0pt}{2.6ex}Benchmark & Base & Standard retrieval & Initial MMKB & MMKB \\
\midrule
\rule{0pt}{2.6ex}MedChain Task~3 & 0.514 & 0.496 & 0.630 & 0.637 \\[2pt]
\arrayrulecolor{rowsep}\specialrule{0.4pt}{0pt}{0pt}\arrayrulecolor{black}
\rule{0pt}{2.6ex}MedThinkVQA & 68.3 & 66.8 & 68.5 & 68.9 \\[2pt]
\arrayrulecolor{rowsep}\specialrule{0.4pt}{0pt}{0pt}\arrayrulecolor{black}
\rule{0pt}{2.6ex}VisuLogic & 45.8 & 46.1 & 49.7 & 50.0 \\[2pt]
\arrayrulecolor{rowsep}\specialrule{0.4pt}{0pt}{0pt}\arrayrulecolor{black}
\rule{0pt}{2.6ex}QCalEval & 72.3 & 75.9 & 83.3 & 84.2 \\[2pt]
\bottomrule
\end{tabular}
\end{adjustbox}
\caption{Absolute scores for the comparison in Figure~\ref{fig:mmkb_retrieval} with GPT-5.6-Sol. MedChain Task~3 reports macro claim recall, and the other benchmarks use the metrics in Section~\ref{sec:experimental_setup}. Standard retrieval and MMKB receive identical retrieved references; Initial MMKB uses only the initial references (Table~\ref{tab:mmkb_sources}) and does not add deployment cases.}
\label{tab:mmkb_retrieval_scores}
\end{table}

\begin{table}[!h]
\centering
\small
\setlength{\tabcolsep}{6pt}
\setlength{\aboverulesep}{0pt}
\setlength{\belowrulesep}{0pt}
\setlength{\arrayrulewidth}{1.2pt}
\renewcommand{\arraystretch}{1.15}
\begin{adjustbox}{scale=\restablescale}
\begin{tabular}{!{\vrule width 0.8pt}c cc cc cc cc!{\vrule width 0.8pt}}
\toprule
\rule{0pt}{17pt}\multirow{2}{*}{MedChain}
&
\multicolumn{2}{c}{\begin{tabular}{@{}c@{}}Nemotron-3-Nano-\\Omni-30B\end{tabular}} &
\multicolumn{2}{c}{GPT-4o-mini} &
\multicolumn{2}{c}{Qwen3.8-27B} &
\multicolumn{2}{c!{\vrule width 0.8pt}}{GPT-5.6-Sol} \\
\cmidrule(lr){2-3}
\cmidrule(lr){4-5}
\cmidrule(lr){6-7}
\cmidrule(lr){8-9}
& Base & Ours
& Base & Ours
& Base & Ours
& Base & Ours \\
\midrule
\rule{0pt}{2.6ex}Task 1 & 52.1 / 0.237 & 57.5 / 0.292 & 58.4 / 0.308 & 58.4 / 0.369 & 60.8 / 0.375 & 58.5 / 0.361 & 57.8 / 0.358 & 65.5 / 0.424 \\[2pt]
\arrayrulecolor{rowsep}\specialrule{0.4pt}{0pt}{0pt}\arrayrulecolor{black}
\rule{0pt}{2.6ex}Task 2 & 0.219 & 0.278 & 0.179 & 0.325 & 0.239 & 0.345 & 0.246 & 0.366 \\[2pt]
\arrayrulecolor{rowsep}\specialrule{0.4pt}{0pt}{0pt}\arrayrulecolor{black}
\rule{0pt}{2.6ex}Task 3 & 0.281 & 0.266 & 0.188 & 0.192 & 0.431 & 0.434 & 0.496 & 0.635 \\[2pt]
\arrayrulecolor{rowsep}\specialrule{0.4pt}{0pt}{0pt}\arrayrulecolor{black}
\rule{0pt}{2.6ex}Task 4 & 0.358 & 0.383 & 0.362 & 0.317 & 0.407 & 0.421 & 0.409 & 0.423 \\[2pt]
\arrayrulecolor{rowsep}\specialrule{0.4pt}{0pt}{0pt}\arrayrulecolor{black}
\rule{0pt}{2.6ex}Task 5 & 0.400 & 0.465 & 0.512 & 0.545 & 0.529 & 0.592 & 0.562 & 0.617 \\[2pt]
\bottomrule
\end{tabular}
\end{adjustbox}
\caption{Per-task MedChain results on unseen cases (Section~\ref{sec:frozen_generalization}). Expertise is learned on the learning split and frozen during evaluation on the test split. Metrics follow Table~\ref{tab:medchain_online}, and their mean gives the MedChain Average in Table~\ref{tab:frozen_results}.}
\label{tab:medchain_frozen}
\end{table}

\begin{table}[!ht]
\centering
\small
\setlength{\tabcolsep}{4pt}
\setlength{\aboverulesep}{0pt}
\setlength{\belowrulesep}{0pt}
\setlength{\arrayrulewidth}{1.2pt}
\renewcommand{\arraystretch}{1.15}
\begin{adjustbox}{scale=\restablescale}
\begin{tabular}{!{\vrule width 0.8pt}c cc cc cc!{\vrule width 0.8pt}}
\toprule
\rule{0pt}{17pt}\multirow{2}{*}{Benchmark}
&
\multicolumn{2}{c}{\begin{tabular}{@{}c@{}}Nemotron-3-Nano-\\Omni-30B\end{tabular}} &
\multicolumn{2}{c}{GPT-4o-mini} &
\multicolumn{2}{c!{\vrule width 0.8pt}}{Qwen3.8-27B} \\
\cmidrule(lr){2-3}
\cmidrule(lr){4-5}
\cmidrule(lr){6-7}
& Base & Transfer
& Base & Transfer
& Base & Transfer \\
\midrule
\rule{0pt}{2.6ex}AgentClinic & 42.1 & 43.0 & 42.5 & 43.5 & 67.3 & 74.3 \\[2pt]
\arrayrulecolor{rowsep}\specialrule{0.4pt}{0pt}{0pt}\arrayrulecolor{black}
\rule{0pt}{2.6ex}MedChain & 0.334 & 0.368 & 0.353 & 0.386 & 0.413 & 0.491 \\[2pt]
\arrayrulecolor{rowsep}\specialrule{0.4pt}{0pt}{0pt}\arrayrulecolor{black}
\rule{0pt}{2.6ex}MedThink-Bench & 42.2 & 44.8 & 37.8 & 39.2 & 64.0 & 67.2 \\[2pt]
\arrayrulecolor{rowsep}\specialrule{0.4pt}{0pt}{0pt}\arrayrulecolor{black}
\rule{0pt}{2.6ex}MedThinkVQA & 25.4 & 26.9 & 23.9 & 28.1 & 48.3 & 50.8 \\[2pt]
\arrayrulecolor{rowsep}\specialrule{0.4pt}{0pt}{0pt}\arrayrulecolor{black}
\rule{0pt}{2.6ex}VisuLogic & 14.4 & 24.3 & 26.6 & 26.3 & 21.5 & 35.8 \\[2pt]
\arrayrulecolor{rowsep}\specialrule{0.4pt}{0pt}{0pt}\arrayrulecolor{black}
\rule{0pt}{2.6ex}QCalEval & 47.3 & 63.2 & 40.9 & 68.2 & 62.9 & 89.8 \\[2pt]
\midrule
\rule{0pt}{2.6ex}MedChain Task~1 & 53.8 / 0.243 & 61.1 / 0.314 & 59.1 / 0.304 & 61.6 / 0.347 & 58.9 / 0.360 & 67.5 / 0.436 \\[2pt]
\arrayrulecolor{rowsep}\specialrule{0.4pt}{0pt}{0pt}\arrayrulecolor{black}
\rule{0pt}{2.6ex}MedChain Task~2 & 0.225 & 0.233 & 0.174 & 0.244 & 0.241 & 0.369 \\[2pt]
\arrayrulecolor{rowsep}\specialrule{0.4pt}{0pt}{0pt}\arrayrulecolor{black}
\rule{0pt}{2.6ex}MedChain Task~3 & 0.238 & 0.261 & 0.190 & 0.218 & 0.410 & 0.463 \\[2pt]
\arrayrulecolor{rowsep}\specialrule{0.4pt}{0pt}{0pt}\arrayrulecolor{black}
\rule{0pt}{2.6ex}MedChain Task~4 & 0.353 & 0.346 & 0.362 & 0.368 & 0.374 & 0.397 \\[2pt]
\arrayrulecolor{rowsep}\specialrule{0.4pt}{0pt}{0pt}\arrayrulecolor{black}
\rule{0pt}{2.6ex}MedChain Task~5 & 0.406 & 0.445 & 0.500 & 0.524 & 0.501 & 0.608 \\[2pt]
\bottomrule
\end{tabular}
\end{adjustbox}
\caption{Absolute scores for zero-step transfer with GPT-5.6-Sol as the source (Figure~\ref{fig:cross_model_transfer}a). Base is each target model's full-stream Base result from Table~\ref{tab:online_results}; Transfer uses the frozen expertise learned with GPT-5.6-Sol without further learning. Each benchmark uses the metric in Section~\ref{sec:experimental_setup}, and MedChain Tasks~1--5 use the metrics in Table~\ref{tab:medchain_online}.}
\label{tab:transfer_sol}
\end{table}

\begin{table}[!ht]
\centering
\small
\setlength{\tabcolsep}{4pt}
\setlength{\aboverulesep}{0pt}
\setlength{\belowrulesep}{0pt}
\setlength{\arrayrulewidth}{1.2pt}
\renewcommand{\arraystretch}{1.15}
\begin{adjustbox}{scale=\restablescale}
\begin{tabular}{!{\vrule width 0.8pt}c cc cc cc!{\vrule width 0.8pt}}
\toprule
\rule{0pt}{17pt}\multirow{2}{*}{Benchmark}
&
\multicolumn{2}{c}{\begin{tabular}{@{}c@{}}Nemotron-3-Nano-\\Omni-30B\end{tabular}} &
\multicolumn{2}{c}{GPT-4o-mini} &
\multicolumn{2}{c!{\vrule width 0.8pt}}{GPT-5.6-Sol} \\
\cmidrule(lr){2-3}
\cmidrule(lr){4-5}
\cmidrule(lr){6-7}
& Base & Transfer
& Base & Transfer
& Base & Transfer \\
\midrule
\rule{0pt}{2.6ex}AgentClinic & 42.1 & 52.8 & 42.5 & 62.6 & 79.0 & 88.3 \\[2pt]
\arrayrulecolor{rowsep}\specialrule{0.4pt}{0pt}{0pt}\arrayrulecolor{black}
\rule{0pt}{2.6ex}MedChain & 0.334 & 0.362 & 0.353 & 0.385 & 0.435 & 0.504 \\[2pt]
\arrayrulecolor{rowsep}\specialrule{0.4pt}{0pt}{0pt}\arrayrulecolor{black}
\rule{0pt}{2.6ex}MedThink-Bench & 42.2 & 47.6 & 37.8 & 41.0 & 68.4 & 77.2 \\[2pt]
\arrayrulecolor{rowsep}\specialrule{0.4pt}{0pt}{0pt}\arrayrulecolor{black}
\rule{0pt}{2.6ex}MedThinkVQA & 25.4 & 25.4 & 23.9 & 25.6 & 68.3 & 67.8 \\[2pt]
\arrayrulecolor{rowsep}\specialrule{0.4pt}{0pt}{0pt}\arrayrulecolor{black}
\rule{0pt}{2.6ex}VisuLogic & 14.4 & 23.4 & 26.6 & 27.2 & 45.8 & 48.7 \\[2pt]
\arrayrulecolor{rowsep}\specialrule{0.4pt}{0pt}{0pt}\arrayrulecolor{black}
\rule{0pt}{2.6ex}QCalEval & 47.3 & 61.0 & 40.9 & 67.8 & 72.3 & 89.1 \\[2pt]
\midrule
\rule{0pt}{2.6ex}MedChain Task~1 & 53.8 / 0.243 & 60.4 / 0.308 & 59.1 / 0.304 & 60.2 / 0.345 & 57.2 / 0.351 & 67.3 / 0.447 \\[2pt]
\arrayrulecolor{rowsep}\specialrule{0.4pt}{0pt}{0pt}\arrayrulecolor{black}
\rule{0pt}{2.6ex}MedChain Task~2 & 0.225 & 0.275 & 0.174 & 0.278 & 0.239 & 0.382 \\[2pt]
\arrayrulecolor{rowsep}\specialrule{0.4pt}{0pt}{0pt}\arrayrulecolor{black}
\rule{0pt}{2.6ex}MedChain Task~3 & 0.238 & 0.223 & 0.190 & 0.192 & 0.514 & 0.509 \\[2pt]
\arrayrulecolor{rowsep}\specialrule{0.4pt}{0pt}{0pt}\arrayrulecolor{black}
\rule{0pt}{2.6ex}MedChain Task~4 & 0.353 & 0.332 & 0.362 & 0.362 & 0.386 & 0.401 \\[2pt]
\arrayrulecolor{rowsep}\specialrule{0.4pt}{0pt}{0pt}\arrayrulecolor{black}
\rule{0pt}{2.6ex}MedChain Task~5 & 0.406 & 0.430 & 0.500 & 0.531 & 0.548 & 0.609 \\[2pt]
\bottomrule
\end{tabular}
\end{adjustbox}
\caption{Absolute scores for zero-step transfer with Qwen3.8-27B as the source (Figure~\ref{fig:cross_model_transfer}b). Base is each target model's full-stream Base result from Table~\ref{tab:online_results}; Transfer uses the frozen expertise learned with Qwen3.8-27B without further learning. Each benchmark uses the metric in Section~\ref{sec:experimental_setup}, and MedChain Tasks~1--5 use the metrics in Table~\ref{tab:medchain_online}.}
\label{tab:transfer_qwen}
\end{table}

\clearpage









\clearpage
\section{Additional Method and Experimental Details}
\label{sec:appendix_method}

\subsection{Learning Settings and Input Limits}

The online batch contains 32 cases, not 32 model calls. The representative buffer also contains 32 cases; four are replaced after each full batch. Every four batches after the first accepted update, we compare the current version with a retained earlier version, using the same MMKB snapshot. A shorter final batch is evaluated but does not trigger optimization. Thus, benchmark size determines the number of update opportunities, while trajectory length and image count determine how much evidence fits into an optimizer request.

The optimizer proposes one complete Skill and at most four knowledge additions per batch. The Skill is limited to 24,000 characters. Knowledge retrieval considers five entries and returns at most two that directly answer the query and satisfy applicability checks, with model-reported confidence at least 0.55. MMKB returns up to three eligible references. Answers and source explanations are excluded from the retrieval representations. GPT-4o-mini performs the knowledge checks; text-embedding-ada-002 and fixed Qwen3-VL-Embedding-2B provide text and multimodal embeddings, respectively.

Input limits do not change the validation batch. MedThinkVQA uses the 16 lowest-scoring trajectories from the 32-case batch to propose an update, but validates on all 32. Its optimizer and the QCalEval optimizer receive current-case images; the VisuLogic optimizer also receives reference images. For MedThinkVQA inference, current images take priority within a limit of 50 images, or 32 for hosted Nemotron, with remaining capacity shared across references. These limits distinguish the experience evaluated from the portion available for proposing an update.

Each comparison uses one execution per case and version. Score comparisons use a tolerance of $10^{-12}$. At a periodic comparison, the current version becomes the retained version if it has a positive mean gain and at least as many wins as losses. A negative mean gain with at least as many losses as wins restores the earlier Skill and Memory; MMKB is unchanged. Otherwise, the current version remains in use. Matching seeds reduces controllable variation but does not guarantee deterministic model responses.

\subsection{Generation Settings}

Table~\ref{tab:generation_settings} separates the base model's generation budget from the optimizer and knowledge checks. These are output limits, not the length of the final answer. A reasoning model may count internal reasoning toward its total output budget. The optimizer receives public explanations and tool traces, not private reasoning traces.

\begin{table}[htbp]
\centering
\small
\begin{tabular}{lp{0.18\linewidth}p{0.41\linewidth}}
\toprule
Role & Output tokens & Reasoning and temperature \\
\midrule
GPT-5.6-Sol base model & 24,576 & High reasoning effort; temperature omitted on routes that do not accept it. \\
GPT-4o-mini base model & 16,384 & Temperature 0; no separate thinking setting. \\
Qwen / Nemotron base models & 24,576 & Temperature 0; thinking budget 8,192 for AgentClinic and MedChain, 16,384 for the four other hosted tasks. \\
Optimizer (GPT-5.6-Sol) & 24,576 & High reasoning effort. \\
Buffer selector (GPT-5.6-Sol) & 4,096 & High reasoning effort. \\
Knowledge checks (GPT-4o-mini) & 4,096 & Temperature 0. \\
\bottomrule
\end{tabular}
\caption{Documented generation settings. The MedChain optimizer and buffer-selector overrides use 65,536 and 16,384 output tokens, respectively; the Skill length limit is unchanged.}
\label{tab:generation_settings}
\end{table}

Knowledge queries are limited to two per case, or two per subquestion in QCalEval. AgentClinic permits up to 20 consultation actions, with knowledge queries counted separately. MedChain history-taking permits up to 10 dialogue turns. These task budgets are distinct from the number of cases in a learning batch.

\subsection{Data and Evaluation}

\begin{table}[htbp]
\centering
\small
\begin{tabular}{lrrr}
\toprule
Benchmark or task pool & Full stream & Learning & Test \\
\midrule
AgentClinic-MedQA & 214 & 149 & 65 \\
MedChain (Tasks 1, 2, 4, 5) & 2,362 & 1,660 & 702 \\
MedChain (Task 3, visual cases) & 693 & 484 & 209 \\
\quad With claim annotations & 561 & 389 & 172 \\
MedThink-Bench & 500 & 350 & 150 \\
MedThinkVQA & 720 & 504 & 216 \\
VisuLogic & 1,000 & 700 & 300 \\
QCalEval & 243 & 170 & 73 \\
\bottomrule
\end{tabular}
\caption{Evaluation pools and fixed learning/test splits. Counts refer to cases. MedChain's visual subset inherits the main case split.}
\label{tab:benchmark_sizes}
\end{table}

All conditions use a fixed case order with seed 42. Splits keep each patient, source group, and quantum experiment intact. One QCalEval case contains all six questions for an experiment, giving 1,458 questions in total. The MedChain split is approximately 70/30 because groups crossing the boundary are kept on the learning side. These are custom splits of public data, not official hidden test sets.

AgentClinic uses final-diagnosis accuracy; MedThink-Bench, MedThinkVQA, and VisuLogic use answer accuracy. QCalEval averages its six task scores within each experiment and then across experiments, using the benchmark's rounding rules and the mean of GPT-5.4 and Gemini-3.1-Pro-Preview grading.

Our MedChain implementation strictly follows the sequential workflow described by \citet{liu2026medchain}: each task receives only the model's outputs from the preceding tasks. In contrast, Table~1 of the original paper reports results in which each task receives the standardized case information. Our scores are therefore not directly comparable to those in that table.

MedChain's Average is the arithmetic mean of six reported metrics across five tasks: first-level referral accuracy, second-level referral IoU, history-taking examination-item IoU, macro claim recall, normalized diagnosis score, and treatment IoU~\citep{liu2026medchain}. The two referral metrics enter separately, rather than being combined before averaging across tasks. Our diagnosis reporting divides the five-point score by five; the internal update-selection score uses $(\text{score}-1)/4$ instead. Claim recall is averaged over annotated cases. Tasks 2 and 5 have 1,984 and 1,921 cases with scorable targets, respectively; their learning/test counts are 1,386/598 and 1,349/572. Missing target annotations do not count as incorrect predictions.

\subsection{Visual Reference Sources}

\begin{table}[htbp]
\centering
\small
\begin{tabular}{lp{0.61\linewidth}}
\toprule
Task & Reference data \\
\midrule
MedChain (visual stage) & 2,722 case--modality records from 2,454 additional cases outside the target pool, with original reports. \\
MedThinkVQA & A fixed 1,024-case subset of official training data, with existing source descriptions. \\
VisuLogic & 4,292 references from official training data, with source answers and available solutions. \\
QCalEval & 82 references (83 images) adapted from public software examples, experimental source data, and published figures; sources are listed in Table~\ref{tab:qcal_reference_sources}. \\
\bottomrule
\end{tabular}
\caption{Initial MMKB libraries. Counts describe stored references, not the number retrieved per case. Original annotations are retained where available; QCalEval sources are adapted into reference questions and answers.}
\label{tab:mmkb_sources}
\end{table}

\paragraph{Preventing answer leakage in MMKB.}
A deployment case is added to MMKB with its source answer only after its whole batch has been answered and scored. Each case can therefore retrieve only cases from earlier batches, and cases in the same batch cannot retrieve one another. A check on batch positions enforces this order, and a run stops with an error if the order is violated. Retrieval also skips the current case, cases from the same source group, and images that are byte-identical, identical after decoding, or near duplicates at thumbnail resolution. Frozen evaluation retains the initial reference library and additions from the learning split only, and transfer retains the source model's final full-stream library; both apply the same exclusions. In online validation, a past case replayed from the buffer cannot retrieve its own entry, including its images and source answer.

\begin{samepage}
\paragraph{QCalEval sources.}
Table~\ref{tab:qcal_reference_sources} identifies the initial references and the versions used to construct them. Repository examples provide data and analysis results; published source data are plotted, and selected paper panels are cropped or rendered. Reference questions and answers are then assembled from these materials. Reported values, computed fits, and explanatory notes are distinguished; the references are not uniformly expert-authored question--answer pairs. The 12 source-data references and one figure reference from Li et al. come from the same study, not independent studies.

\end{samepage}

\begin{table}[htbp]
\centering
\small
\begin{tabular}{p{0.30\linewidth}rp{0.54\linewidth}}
\toprule
Source & Count & Material used \\
\midrule
\href{https://github.com/quantify-os/quantify-core/tree/344c8bc49ef9be9f66be5eef3c68ac609ee6c465}{Quantify-core (344c8bc)} & 34 & Example data and analysis code; references use raw data or recomputed fits. \\
\href{https://github.com/qiboteam/qibocal/tree/e0ba08ab119de9f25ab1cc97da73853bfc28e0d2}{Qibocal (e0ba08a)} & 26 & Calibration test data, existing fits, ZZ data, and a documentation figure. \\
\href{https://github.com/QuTech-Delft/qtt/tree/77a5a86f37a822e57d6ca10f093526a1134af819}{QTT (77a5a86)} & 1 & Barrier-gate pinch-off scan (B8.dat), with metadata and notebook. \\
\href{https://doi.org/10.6084/m9.figshare.28193546.v1}{Li et al., source data (v1)} & 12 & Plots generated from public measurement spreadsheets, with paper descriptions and reported parameters. \\
\midrule
\citet{li2025fiber} & 1 & Fig.~1c: atom-array rearrangement. \\
\citet{padilla2025large} & 2 & \href{https://arxiv.org/abs/2506.01180v1}{arXiv v1}, Figs.~2a and 5b: magneto-optical-trap loading and accumulation. \\
\citet{senoo2026high} & 4 & \href{https://arxiv.org/abs/2506.13632v2}{arXiv v2 (2025)}, Extended Data Figs.~4b--d, 4e--f, and 7c--d; main Fig.~3b--d: Ramsey, Rabi, gate benchmarking, and interaction spectra. \\
\citet{ali2024reducing} & 2 & Figs.~2a--c and 10a--d: two-state and three-state readout distributions. \\
\bottomrule
\end{tabular}
\caption{Sources of the 82 initial QCalEval MMKB references. Links identify repository revisions, dataset versions, or the paper versions used. The last four rows contribute nine references with ten images.}
\label{tab:qcal_reference_sources}
\end{table}

\clearpage
\section{Prompts and Message Construction}
\label{sec:appendix_prompts}

We report the fixed prompts that govern Skill execution, update proposals, knowledge checks, and visual reference use. The boxes reproduce the documented prompt text; double braces mark values filled at runtime, not text sent literally to the model. Role labels and box headings are explanatory.

A learned Skill is an output of optimization, not a fixed prompt shared by all runs. Skill and Knowledge Memory start empty; image tasks also use the initial MMKB libraries in Table~\ref{tab:mmkb_sources}. The base model receives the current Skill, permitted tool responses, and any retrieved references. It does not receive the current case's answer or score. The optimizer receives completed trajectories and their released feedback. Graders have access to the scoring targets but do not revise the model's answer.

\subsection{Executing the Current Skill}

The task instructions define the allowed inputs, tools, and answer format. The following protocol specifies how the current Skill is used within that task. AgentClinic retains its native multi-turn messages; the other task adapters insert the Skill into their task-specific system message.

\Needspace{6\baselineskip}
\noindent\textbf{Skill execution protocol}\par
\begin{lstlisting}[style=paperprompt]
Generic Skill execution protocol:
1. Treat the learned Skill as executable policy, not as facts about the input.
2. Execute `CORE` once in written order; treat unlabelled procedural instructions as `CORE`.
3. At each decision point, evaluate each `WHEN` against the current input and current execution state. Execute only the matching `DO`; an unmatched rule has no effect.
4. After an action changes the execution state, re-evaluate the remaining rules.
5. End a rule when its `STOP` condition holds, and do not repeat it unless the Skill explicitly says to do so.
6. The task contract and direct input evidence override conflicting Skill text.
Keep rule selection and execution bookkeeping internal; return only the task's required output.
\end{lstlisting}

\subsection{Proposing Updates}

The optimizer receives the current Skill and knowledge catalog, scored trajectories, and feedback from earlier comparisons when available. 

\Needspace{6\baselineskip}
\noindent\textbf{Optimizer: system message}\par
\begin{lstlisting}[style=paperprompt]
You optimize a reusable Skill and propose independently verified Knowledge Memory candidates from scored trajectories. Return one json object only; never answer an individual training case.
\end{lstlisting}
\Needspace{10\baselineskip}
\noindent\textbf{Optimizer: user message}\par
\begin{lstlisting}[style=paperprompt]
Task contract:
{{TASK_CONTRACT}}

Optimizer branch: online

Current complete Skill:
<skill>
{{CURRENT_SKILL_OR_EMPTY_MARKER}}
</skill>

Skill document convention:
A Skill document has lightweight execution semantics:
- `CORE` contains instructions that run once, in written order, for every input.
- An optional conditional rule has a stable rule id and `WHEN`, `DO`, and `STOP`.
- `WHEN` is checked only against the current input and current execution state.
- `DO` is executed only when its `WHEN` is satisfied.
- `STOP` states the observable condition that ends that rule.
- Unlabelled procedural instructions are treated as `CORE` for compatibility.

The convention defines how a Skill is executed. It does not prescribe a task workflow or require any fixed number of sections or conditional rules.

Current Knowledge Memory:
{{MEMORY_CATALOG_JSON}}

The trajectory records below contain outcomes from the task's deterministic or replayable verifier. Treat `primary_score` and per-item verifier labels as ground truth feedback: larger primary_score is better. Infer the complete executable Skill most likely to improve future task scores. You may learn from observed behavior or propose a new testable strategy from the task objective. You may retain, modify, delete, or rewrite the current Skill.

{{TASK_SCOPE_POLICY}}
Return the complete Skill file, no more than 24000 total rendered characters (no rule-count limits).
An empty candidate means keep the current complete Skill unchanged.

Knowledge Memory stores WHAT is true, never an instruction. Optionally propose up to 4 atomic declarative knowledge candidates. Each candidate must state the question it answers, minimal applicability, one-based supporting trajectory indices, and exact supporting quotes. The runtime independently verifies them. Two independent trajectory records may establish the same reusable relation; a candidate supported by only one record requires a directly supporting trusted Web excerpt. {{TASK_MEMORY_POLICY}}
It is valid to return no memory candidates.

{{PREVIOUS_PROPOSAL_FEEDBACK_SECTION}}
{{LONGITUDINAL_SECTION}}
{{SHARED_TEXT_DEFINITIONS}}

Return json only with exactly these fields:
{
  "analysis": "task-level score diagnosis and strategy",
  "candidate_skill": "the complete Skill file",
  "memory_candidates": [{
      "question": "knowledge question",
      "knowledge": "one declarative relation",
      "applies_to": "minimal conditions",
      "supporting_record_indices": [1],
      "supporting_quotes": ["exact quote"]
    }],
  "rationale": "why this Skill may improve future scores"
}

Verified trajectories:
{{VERIFIED_TRAJECTORIES_JSON}}
\end{lstlisting}

\paragraph{Task-specific and optional inputs.}
The task contract specifies the objective and valid actions. The scope and memory policies state what can be reused in that domain: procedures belong in Skill, while supported declarative knowledge belongs in Knowledge Memory. Their four profiles cover the clinical text/workflow tasks, MedThinkVQA, VisuLogic, and QCalEval. Completed trajectories supply public actions, explanations, tool responses, and verifier feedback. Optional sections supply the previous proposal's evaluation, periodic comparisons, and definitions for repeated text within the same request. Absent sections are omitted. Repeated text is shared exactly, not summarized.

\subsection{Checking and Retrieving Knowledge}

A retrieved entry must answer the model's explicit information need and meet its applicability conditions. A shared topic alone is not enough. The selector may return no entry.

\Needspace{6\baselineskip}
\noindent\textbf{Knowledge selection: system message}\par
\begin{lstlisting}[style=paperprompt]
Select only directly applicable declarative knowledge. Return json only.
\end{lstlisting}
\Needspace{6\baselineskip}
\noindent\textbf{Knowledge selection: user message}\par
\begin{lstlisting}[style=paperprompt]
Explicit knowledge need:
{{INFORMATION_NEED_JSON}}

Candidate declarative knowledge:
{{TOP5_KNOWLEDGE_JSON}}

Select zero to 2 nonredundant candidates. A candidate is eligible only if it directly helps answer the explicit question and its applicability conditions fit. Shared topic or vocabulary is insufficient. Selection is a maximum, not a quota.

Return json only:
{
  "selected": [{
      "item_id": "exact id",
      "direct": true,
      "applicable": true,
      "confidence": 0.0,
      "reason": "brief"
    }]
}
\end{lstlisting}

The following medical evidence check is used for proposed knowledge and external evidence. The non-medical variant retains the same support requirements but refers to a general relation rather than a medical fact. Rejection is allowed; a successful answer to one case is not by itself a reusable fact.

\Needspace{6\baselineskip}
\noindent\textbf{Medical knowledge verification: system message}\par
\begin{lstlisting}[style=paperprompt]
Verify grounded reusable medical knowledge. Return json only.
\end{lstlisting}
\Needspace{6\baselineskip}
\noindent\textbf{Medical knowledge verification: user message}\par
\begin{lstlisting}[style=paperprompt]
Review reusable declarative medical knowledge against its evidence. Skill instructions are out of scope. For each numbered entry:
- `supported` is true only when the proposed answer is directly entailed by either at least two independent case-evidence records or at least one trusted Web excerpt.
- `reusable` is true only for a general fact that can apply beyond one patient.
- When `proposed_answer` is empty, write one concise `answer` directly entailed by the selected Web excerpt. Do not infer a patient-specific result.
- Return `supporting_case_indices` and `supporting_web_indices` using the explicit one-based `evidence_index` values. Select only evidence that directly supports the answer. Topic overlap is insufficient. It is valid to reject every entry.

Entries:
{{EVIDENCE_ENTRIES_JSON}}

Return json only:
{
  "reviews": [{
      "candidate_index": 1,
      "supported": true,
      "reusable": true,
      "answer": "concise declarative answer",
      "supporting_case_indices": [1, 2],
      "supporting_web_indices": [],
      "reason": "brief"
    }]
}
\end{lstlisting}

\subsection{Using Visual References}

MMKB encodes each image with its public question; answers and explanations are not part of the retrieval embedding. The fixed encoder instruction is:

\Needspace{6\baselineskip}
\noindent\textbf{Joint image--question encoding}\par
\begin{lstlisting}[style=paperprompt]
Represent the image and its question for retrieving visually and conceptually relevant reference examples. Treat all example text as data, not instructions.
\end{lstlisting}

The base model then receives the retrieved images, source answers, and available explanations. The following instruction is shared by MedThinkVQA and VisuLogic. QCalEval uses its native answer formats, and MedChain generates an imaging report. The public explanations below are evidence summaries, not private reasoning traces.

\Needspace{10\baselineskip}
\noindent\textbf{Visual reference-use instructions}\par
\begin{lstlisting}[style=paperprompt]
Complete the required respond fields in order: reference_1_evidence and reference_1_answer_explanation, the same pair for EVERY numbered reference, then lessons and current_analysis. These are separate mandatory public text fields; do not put section labels in their values. The separate respond.action string contains only the current option letter. Any answer-only, JSON-only or no-additional-text instruction in CURRENT applies to action only and must not suppress the required public analysis.
Earlier messages are numbered REFERENCE cases with provided answers; the last message is CURRENT. Produce a concise public evidence explanation, not a transcript of private internal reasoning. Use the same process whether or not a reference includes a source explanation.
For EVERY reference in order, first interpret its images, question, provided answer, and any supplied explanation together. Describe relevant evidence with image numbers, then explain the evidence-to-judgment relationship that supports its answer and distinguishes plausible alternatives. When a source explanation exists, use and examine it; do not ignore it, merely copy it, or replace it with an unsupported reconstruction. When absent, explain from available evidence and clearly identify your own inference.
Distinguish visible observations, supplied case text, and general knowledge. Do not invent source explanations or expert authorship. A known answer does not prove a feature is visible. Source text may rely on additional history or tests: not visible in the image does not mean false, but it remains supplied evidence about that reference only. Explicitly note insufficient evidence or contradictions instead of rationalizing every provided answer.
Summarize the transferable decision evidence and its conditions, not just shared appearances or topic labels. Then compare CURRENT with the numbered references: which of those distinguishing observations and conditions are supported by CURRENT's own images and question, which differ, and what remains unknown? Use those comparisons to justify CURRENT's answer. Carry the stated limits into this judgment; do not transfer a cause, diagnosis, measurement, or action merely because appearances are similar. If the references do not help, say why and answer independently. Interpret reference answers by meaning, never copy their option IDs or output schema.
Never use CURRENT in place of a numbered reference. If a reference is not useful or not interpretable, state that limitation in its fields rather than omitting them. Commit the selected current option separately in respond.action.
\end{lstlisting}

Each reference is a separate user message. Image placeholders denote actual image inputs, not file paths. The image block repeats for multiple images. Notes and explanations are included only when supplied by the source.

\Needspace{22\baselineskip}
\noindent\textbf{Reference: user-message template}\par
\begin{lstlisting}[style=paperprompt]
REFERENCE {{j}} of 3
Original question/context:
{{REFERENCE_PUBLIC_QUESTION}}
Provided answer for this reference:
{{REFERENCE_SOURCE_ANSWER}}

Reference {{j}} source type: {{SOURCE_KIND}}.
Reference {{j}}, image {{image_index}}:
{{IMAGE_PART}}
Provided image note for reference {{j}}, image {{image_index}} (supplied text; not independently verified):
{{SOURCE_IMAGE_NOTE}}

Provided explanation for reference {{j}} (supplied text; not independently verified):
{{SOURCE_EXPLANATION}}
\end{lstlisting}
\Needspace{6\baselineskip}
\noindent\textbf{Current case: user-message template}\par
\begin{lstlisting}[style=paperprompt]
CURRENT
{{CURRENT_PUBLIC_QUESTION}}
Current image {{image_index}}:
{{CURRENT_IMAGE_PART}}
\end{lstlisting}

Source types distinguish expert-authored, model-generated, dataset-provided, and unspecified annotations. Dataset-provided or unspecified text does not establish expert authorship. The complete response schema requires evidence and answer explanations for each reference, transferable lessons, analysis of the current input, and a separate final answer.

\subsection{Feedback and Representative Cases}

Validation is performed by the comparisons in Section~\ref{sec:joint_validation}, not by asking the optimizer to approve its own proposal. The feedback message distinguishes a rejected proposal from an update that was actually executed.

\Needspace{6\baselineskip}
\noindent\textbf{Previous-proposal feedback inserted into the optimizer message}\par
\begin{lstlisting}[style=paperprompt]
Previously evaluated proposals relevant to the current checkpoint:
The same cases and seeds were run with the incumbent and candidate. Use their final outcomes and stage attribution to determine which parts helped and which parts hurt. A whole candidate may contain both useful and harmful components. Revise, reuse, or discard ideas according to the causal evidence and gate outcome.
When `rejection_stage` is `candidate_validation`, the candidate was not executed; obey its exact validation error and return a compliant revision.
When feedback has separate `validation`, `tested_checkpoint`, and `effect` fields, read them separately: an invalid proposed Skill was not run. A tested checkpoint may contain the old Skill plus independently accepted Knowledge Memory. Its gate effect describes that actual checkpoint, not the rejected new Skill. A resolved length validation describes a repaired Skill; use the actual tested Skill and do not repeat an already resolved repair.
{{FEEDBACK_JSON}}
\end{lstlisting}

The buffer selector sees only the task objective and the current and incoming question texts, not scores or reference answers. It selects four one-for-one replacements to improve problem coverage.

\end{document}